\documentclass[11pt]{article}

\usepackage[preprint]{acl}

\usepackage{times}
\usepackage{latexsym}

\usepackage[T1]{fontenc}
\usepackage[utf8]{inputenc}

\usepackage{microtype}
\usepackage{inconsolata}
\usepackage{graphicx}

\usepackage{enumitem}
\usepackage{svg}
\usepackage{makecell}
\usepackage{tabularx}
\usepackage{longtable}
\usepackage{pifont}
\usepackage{caption}
\usepackage{booktabs}
\usepackage{multirow}
\usepackage{colortbl}
\usepackage{amsmath}
\usepackage{float}
\usepackage{balance}
\usepackage{siunitx}

\usepackage{amsfonts}
\usepackage{amssymb}

\usepackage[english, status=draft, margin=false, inline=true]{fixme}
\fxusetheme{color}
\FXRegisterAuthor{jf}{ajf}{\color{red}JF}
\FXRegisterAuthor{gs}{ags}{\color{blue}GS}
\FXRegisterAuthor{gc}{agc}{\color{purple}GC}

\usepackage[linesnumbered, ruled, vlined]{algorithm2e}

\usepackage[table,xcdraw]{xcolor}
\definecolor{zhawblue}{rgb}{0.00, 0.39, 0.65}
\definecolor{codegreen}{rgb}{0,0.6,0}
\definecolor{codegray}{rgb}{0.5,0.5,0.5}
\definecolor{codepurple}{rgb}{0.58,0,0.82}
\definecolor{codebackground}{rgb}{0.93,0.94,0.95}
\definecolor{lightbrown}{RGB}{150, 75, 0}

\usepackage{listings}
\lstdefinestyle{jsonstyle}{
  language=json,
  basicstyle=\ttfamily\footnotesize,
  breaklines=true,
  columns=flexible,
  frame=single,
  backgroundcolor=\color{gray!5},
  keywordstyle=\color{blue},
  stringstyle=\color{teal},
  commentstyle=\color{gray},
  showstringspaces=false,
}

\lstdefinestyle{toolstyle}{
  language=Python,
  basicstyle=\ttfamily\footnotesize,
  breaklines=true,
  backgroundcolor=\color{gray!5},
  showstringspaces=false,
  columns=fullflexible
}

\usepackage{listings}
\usepackage[strings]{underscore}
\usepackage[cache=false]{minted}
\usepackage[most]{tcolorbox} %
\tcbuselibrary{listings, breakable, minted, skins, listingsutf8}

\usepackage[most]{tcolorbox}

\tcbset{
  myjsonstyle/.style={
    enhanced,
    colback=gray!5,
    colframe=black!70,
    fonttitle=\bfseries,
    boxrule=0.5pt,
    arc=2pt,
    outer arc=1pt,
    top=2pt, bottom=2pt, left=2pt, right=2pt,
    listing only,
    listing options={
      language=json,
      basicstyle=\ttfamily\footnotesize,
      breaklines=true,
      columns=fullflexible,
      keepspaces=true,
      showstringspaces=false,
      morecomment=[l]{//}
    }
  }
}

\lstdefinelanguage{json}{
  morestring=[b]",
  morecomment=[l]{//},
  morekeywords={true,false,null},
  sensitive=false
}
\definecolor{lightbrown}{RGB}{150, 75, 0}

\lstdefinestyle{mystyle}{
    backgroundcolor=\color{codebackground},
    commentstyle=\color{codegreen},
    keywordstyle=\color{magenta},
    numberstyle=\footnotesize\color{codegray},
    stringstyle=\color{lightbrown},
    basicstyle=\ttfamily\footnotesize,
    breakatwhitespace=false,
    breaklines=true,
    keepspaces=true,
    numbers=left,
    numbersep=5pt,
    showspaces=false,
    showstringspaces=false,
    showtabs=false,
    tabsize=4
}
\tcbset{
  myjsonstyle/.style={
    enhanced,
    colback=gray!5,
    colframe=black!70,
    fonttitle=\bfseries,
    boxrule=0.5pt,
    arc=2pt,
    outer arc=1pt,
    top=2pt, bottom=2pt, left=2pt, right=2pt,
    listing only,
    listing options={
      language=json,
      basicstyle=\ttfamily\footnotesize,
      breaklines=true,
      columns=fullflexible,
      keepspaces=true,
      showstringspaces=false,
      morecomment=[l]{//}
    }
  }
}

\lstdefinestyle{jsonstyle}{
  language=json,
  basicstyle=\ttfamily\footnotesize,
  breaklines=true,
  columns=flexible,
  frame=single,
  backgroundcolor=\color{gray!5},
  keywordstyle=\color{blue},
  stringstyle=\color{teal},
  commentstyle=\color{gray},
  showstringspaces=false,
}

\lstdefinestyle{toolstyle}{
  language=Python,
  basicstyle=\ttfamily\footnotesize,
  breaklines=true,
  backgroundcolor=\color{gray!5},
  showstringspaces=false,
  columns=fullflexible
}

\renewcommand{\thelisting}{\arabic{listing}}

\newcommand*\circled[1]{\tikz[baseline=(char.base)]{
            \node[shape=circle,draw,inner sep=.6pt] (char) {#1};}}

\newcommand{\rmspace}{\vspace{-1ex}}

\title{AgenticIE: An Adaptive Agent for Information Extraction from Complex Regulatory Documents}

\author{
Gaye Colakoglu$^{1, 2}$  \quad Gürkan Solmaz$^{2}$ \quad Jonathan Fürst$^{1}$
\\
$^{1}$Zurich University of Applied Sciences, Switzerland \\
$^{2}$NEC Laboratories Europe, Heidelberg, Germany\\
\texttt{colgay01@students.zhaw.ch}\\
\texttt{guerkan.solmaz@neclab.eu}\\ \texttt{jonathan.fuerst@zhaw.ch}
}

\begin{document}
\maketitle
\begin{abstract}
Declaration of Performance (DoP) documents, mandated by EU regulation, specify characteristics of construction products, such as fire resistance and insulation. While this information is essential for quality control and reducing carbon footprints, it is not easily machine readable. Despite content requirements, DoPs exhibit significant variation in layout, schema, and format, further complicated by their multilingual nature. In this work, we propose DoP Key Information Extraction (KIE) and Question Answering (QA) as new NLP challenges. To address this challenge, we design a domain-specific AgenticIE system based on a planner-executor-corresponder pattern. For evaluation, we introduce a high-density, expert-annotated dataset of complex, multi-page regulatory documents in English and German. Unlike standard IE datasets (e.g., FUNSD, CORD) with sparse annotations, our dataset contains over 15K annotated entities, averaging over 190 annotations per document. Our agentic system outperforms static and multimodal LLM baselines, achieving Exact Match (EM) scores of 0.396 vs. 0.342 (GPT-4o, +16\%) and 0.314 (GPT-4o-V, +26\%) across the KIE and QA tasks. Our experimental analysis validates the benefits of the agentic system, as well as the challenging nature of our new DoP dataset.

\end{abstract}

\section{Introduction}
\label{sec:introduction}

Although EU regulations specify the required content, they do not define a specific data model or format. DoPs can be scanned or digital, structured or semi-structured, and multilingual. Layouts vary by manufacturer, with key information spread across tables, free-form text, and hybrid formats.

\begin{figure}[t]
    \centering
    \includegraphics[width=\linewidth]{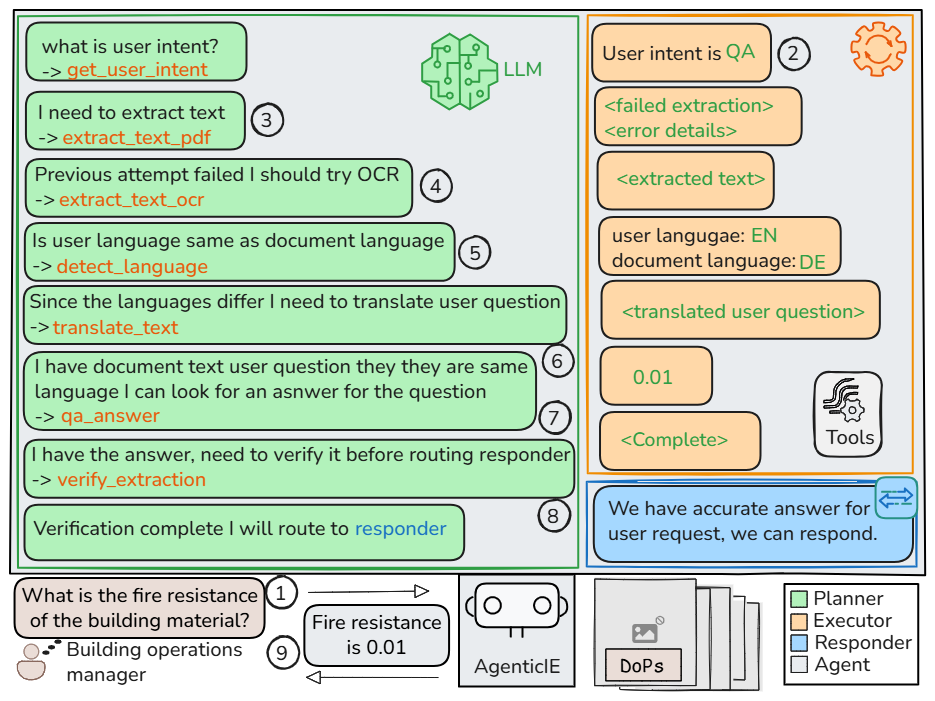}
\caption{AgenticIE: The user's request about the characteristic of a building material triggers a workflow with internal planner, executor, and responder coordination.
}
    \label{fig:system-workflow}
    \rmspace
\end{figure}

To make the contained information in DoP documents machine and human accessible, we propose DoP Information Extraction (IE) as a new application to the NLP community.
IE from DoPs reduces manual work, improves digital compliance processes and enables automated quality control in building construction (e.g., through Building Information Modeling, BIM~\cite{garrido2022interlinking}).

Current IE methods for layout-rich documents usually build on custom or general transformer-based models~\cite{colakoglu-etal-2025-problem}. They can be divided into text-only methods and multi-modal methods.
Still, they often struggle with noisy scans, inconsistent layouts, and multiple languages~\cite{xu2020layoutlm, kim2021donut, sarkhel2021improving, perot2023lmdx}.
Recently, Agentic AI has been proposed to tackle IE with a diverse document structure~\cite{landingai2025agentic}. Unlike static single-shot text- or vision-only systems, agentic systems can track decisions, recover from errors, and adapt tools dynamically~\cite{nooralahzadeh2024explainable}.
Our intuition is that these characteristics make them a better fit for IE tasks from complex, real-world documents.

\textbf{AgenticIE Method.} First, we propose a domain-specific agentic system for DoP documents that infers user intent, detects modality (scanned or digital), and selects tools via a planner-executor-responder loop. Inspired by tool-augmented agents~\cite{schick2023toolformer}, modular planners~\cite{liu2023agentbench}, and structured frameworks~\cite{wu2024autogen}, it maintains state and adapts to user goals and document structure for improved robustness.
To show the working and benefits of AgenticIE, let us consider an example (Figure~\ref{fig:system-workflow}): \circled{1} The user provides an input to the system: ``What is the fire resistance rating in this DoP?''; \circled{2} The LLM determines user intent and predicts a QA task; \circled{3} The agent calls a tool to extract the text, which fails immediately as the PDF is image-based; \circled{4} Based on the failed extraction the agent chooses to use an OCR tool to extract the text from the image which succeeds; \circled{5} A language detector tool is called to detect the user and document language which predicts the user question to be in German and the document English; \circled{6} The agent decides to translate the question to English to match the document language; \circled{7} Now with question and document text ready, the QA tool is called to retrieve the fire resistance: 0.01; \circled{8} The answer is verified and \circled{9} returned in user language to the user.

\textbf{DoP Dataset.} Second, we create a new resource of multi-page documents (174 pages) of which we annotate more than 15K key-value and question answer pairs in English and German. 
Our agentic system outperforms a static LLM and Multimodal LLM baseline, achieving Exact Match (EM) scores of 0.396 vs. 0.342 (GPT-4o, +16\%) and 0.314 (GPT-4o-V, +26\%) across Key Information Extraction (KIE) and QA tasks.

\noindent In summary, our main contributions are:

\begin{itemize}[leftmargin=*,noitemsep,topsep=0pt]

\item We release the first, multilingual (English and German) dataset of DoP documents with overall more than 15K human-labeled KIE and QA annotations (Section~\ref{sec:data_construction}).

\item We design an agentic system that adapts to user intent and document modality for KIE and QA tasks on DoP documents (Section~\ref{sec:agentic_dop_understanding}).

\item We extensively evaluate KIE and QA against GPT-4o text and vision baselines (schema and value validity), demonstrating improved generalization across layout and language variations (Section~\ref{sec:experimental_setup} and \ref{sec:evaluation}).

\end{itemize}

The AgenticIE code repository and the annotated DoP dataset are available as supplementary files and will be released online for usage and reproduction.

\section{DoP IE Problem}
\label{sec:dop-problem}

\begin{figure*}[ht]
    \centering
    \includegraphics[width=1.0\linewidth]{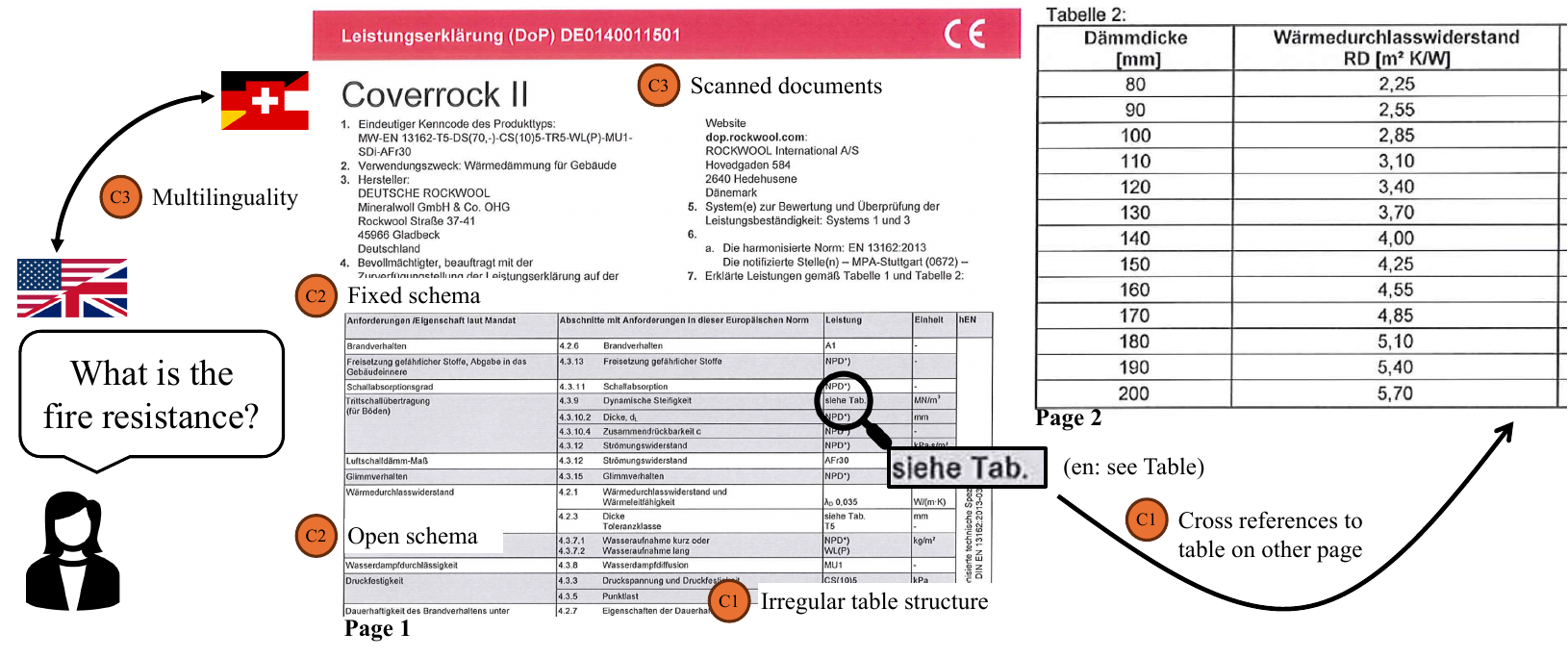}
\caption{German DoP Example with IE challenges highlighted (\textbf{[C1], [C2], [C3]}). First, text needs to be extracted from the scanned document. Second, nested \textit{key:value} information needs to be extracted according to a given schema, including complex cross-referenced data on different pages. User and document language do not match.}
    \label{fig:dop-example}
    \rmspace
\end{figure*}

IE from DoP documents presents unique difficulties that are distinct from standard visually rich document understanding (VRDU). We identify three key challenges depicted through an example in Figure~\ref{fig:dop-example}:

\paragraph{[C1] Irregular Table Structure and Cross-References.} 
DoPs exhibit high layout variation, mixing free-form text with complex, borderless tables spanning multiple pages. A critical challenge is resolving implicit cross-references, where values (e.g., product codes) link to specifications in disjoint tables (Figure~\ref{fig:dop-example}). Standard OCR models often fail to capture these non-local dependencies, resulting in fragmented extractions.

\paragraph{[C2] Hybrid Schema: Fixed vs. Open Structure.} 
DoP extraction requires handling hybrid schemas. While standard metadata (e.g., \textit{Manufacturer}, \textit{DoP Number}) follow a \textbf{Fixed Schema} where keys are known a priori, technical specifications (e.g., under \textit{Declared Performance}) follow an \textbf{Open Schema}. In these nested sections, the keys (sub-properties) are dynamic, unknown beforehand, and vary significantly between product types. Systems cannot rely on static templates, but they must dynamically discover and structure these open-ended attributes.

\paragraph{[C3] Multilinguality and Scanned Documents.} 
DoPs are in different languages, often creating asymmetry, where the user's query language ($L_{user}$) differs from the document's ($L_{doc}$). Systems must perform on-the-fly semantic alignment to map intents (e.g., English ``Fire Resistance'') to native document values (e.g., German ``Brandverhalten'').

\subsection{Task Definition}

We define DoP IE with its two related sub-tasks: \textbf{Key Information Extraction (KIE)} and \textbf{Question Answering (QA)}.
Let $\mathcal{D}$ be a document with language $L_{doc}$ and $L_{user}$ be the user's target language.

\paragraph{Sub-Task 1: Hybrid KIE}
Given $\mathcal{D}$ and a target schema in $L_{user}$, the goal is to extract a structured object $O$. The extraction involves two types of targets:
1. \textbf{Fixed Keys:} Predefined fields (e.g., metadata) provided explicitly to the model.
2. \textbf{Open Keys:} Dynamic fields (e.g., \textit{Declared Performance}) where the model must autonomously discover a nested substructure $v = \{subk_1: v'_1, \dots, subk_m: v'_m\}$ without prior knowledge of the sub-keys.
The system must preserve this hierarchy and return content in $L_{user}$, translating if $L_{doc} \neq L_{user}$.

\paragraph{Sub-Task 2: Multilingual QA.}
Given $\mathcal{D}$ and question $Q$ in $L_{user}$, generate answer $A$ in $L_{user}$. $A$ must be grounded in $\mathcal{D}$, resolving cross-references and tabular structures.

\paragraph{Precision Requirement.}
For both sub-tasks, EM is essential. Unlike general-purpose retrieval where semantic similarity often suffices, the extracted information is used directly to select materials based on specific characteristics (e.g., fire resistance class, thermal conductivity). Consequently, precise extraction of values and units is critical, as approximations could lead to the selection of non-compliant materials.

\section{Dataset Construction}
\label{sec:data_construction}

We introduce a corpus of DoP documents, mandated by the EU Construction Products Regulations (EU No 305/2011, EU No 574/2014)~\cite{eu574_2014}. The documents were collected via web searches and construction repositories (e.g., BuildUp~\cite{buildup}) and selected to capture high variability in layout, ranging from simple single-column text to complex, multi-page tabular structures.

\subsection{KIE Annotation}
\label{sec:target_fields}

We annotated 12 key fields essential for regulatory compliance (see~\ref{appendix:12-target-keys}). Consistent with the \textbf{Hybrid Schema} challenge (Section~\ref{sec:dop-problem}), our annotation strategy distinguishes between two data types:
\begin{itemize}[leftmargin=*]
    \item \textbf{Fixed Keys:} Metadata fields (e.g., \textit{Manufacturer, Declaration number}) with flat values.
    \item \textbf{Open Keys:} Complex fields like \textit{Declared Performance} that follow an open schema. These are stored as nested JSON objects to preserve their hierarchical structure and semantic relationships. Their schema differs between products and manufacturers.
\end{itemize}

Listing~\ref{lst:kie_annotation_example} shows annotations for a German document, capturing both the English and German keys alongside their nested values (full version~\ref{appendix:kie-annotation}).

\refstepcounter{listing}
\begin{tcolorbox}[
  title=Listing \thelisting: KIE Annotation Example, center title, center, breakable,
  minted language=json,
  colback=orange!5, colframe=brown, boxrule=0.4pt,
  boxsep=0pt, top=2pt, bottom=2pt, left=5pt, right=5pt
]
\label{lst:kie_annotation_example}
\footnotesize
\{
  "doc_name": "doc001.pdf", "doc_lang": "de",
  "key_en": "Declared Performance", "key_de": "Erklaerte Leistung",
  "value_en": {...}, "value_de": {...}
\}
\end{tcolorbox}

\subsection{QA Generation}
\label{sec:qa_data_generation}
We programmatically generate natural language questions from the manual KIE annotations using a template-based approach, where we developed 22 distinct question templates (11 per language) organized into three categories: hierarchical path-based templates that use contextual information from the key structure, custom question mappings for domain-specific terminology, and entity-type templates that adapt based on field semantics. Template selection follows a hierarchical decision tree based on key patterns and semantic context (see~\ref{appendix:qa-templates}).

\paragraph{Template-Based Generation.}
We employ semantic role templates to generate fluent questions in both English and German in the following categories: (1) Factual: \textit{``What is the Declaration Number?''};
(2) Entity-Centric: \textit{``Who is the Manufacturer?''};   (3) Regulatory: \textit{``Which AVCP system applies according to...?''}

\paragraph{Nested Structures.}
For open-schema keys, we utilize path-aware templates to target specific sub-properties. For example, a value nested under \textit{Declared Performance} $\rightarrow$ \textit{Reaction to Fire} generates the query: \textit{"What is the value of 'Declared Performance/Reaction to Fire'?"}. Similarly, signature extraction is targeted via contextual prompts like \textit{"Who signed the declaration?"}.

Listing~\ref{lst:qa_json} shows a generated QA entry; the full annotation for one document is in Appendix~\ref{appendix:qa-annotation}.

\refstepcounter{listing}
\begin{tcolorbox}[
  title=Listing \thelisting: QA Generation Example, center title, center, breakable,
  colback=orange!5, colframe=brown, boxrule=0.4pt,
  boxsep=0pt, top=2pt, bottom=2pt, left=5pt, right=5pt
]
\label{lst:qa_json}
\footnotesize
\{
  "doc\_name": "doc001.pdf", "user\_language": "en",
  "parent\_key": "Manufacturer", "key": "Manufacturer",
  "value": "Company X, Switzerland",
  "question": "Who is the Manufacturer?"
\}
\end{tcolorbox}

\subsection{Dataset Statistics}
Our annotated dataset comprises 80 DoP PDF documents, 50 in German and 30 in English. Of these, 65 are text-based (digitally generated), whereas the remaining files are scanned, image-based PDFs.
The dataset consists of 174 pages with documents spanning from 1 to 16 pages.
For KIE, we annotated 12 key fields per document in both languages, resulting in 1,920 annotations. If keys contain nested data, we expand that structure and also manually annotate it. This results in 5,765 additional annotations. For QA, we generate an entry for each key–language pair in the key fields (1,920 QA pairs) and nested data (5,727 QA pairs).
Overall, this results in 15,332 annotations (Table~\ref{tab:kie_qa_statistics}).

\begin{table}[ht]
\centering
\caption{Summary of the KIE and QA annotations.}
\label{tab:kie_qa_statistics}
\footnotesize
\begin{tabular}{@{}lrr@{}}
\toprule
\textbf{Category} & \textbf{KIE Targets} & \textbf{QA Pairs} \\
\midrule
Fixed Schema Keys & 1,920 & 1,920 \\
Open Schema Keys (Nested) & \textbf{5,765} & \textbf{5,727} \\
\midrule
\textbf{Total Annotations} & \multicolumn{2}{c}{\textbf{15,332}} \\
\bottomrule
\end{tabular}
\rmspace
\end{table}

\section{Agentic DoP Understanding}
\label{sec:agentic_dop_understanding}

\begin{figure}[ht]
    \centering
    \includegraphics[width=\columnwidth]{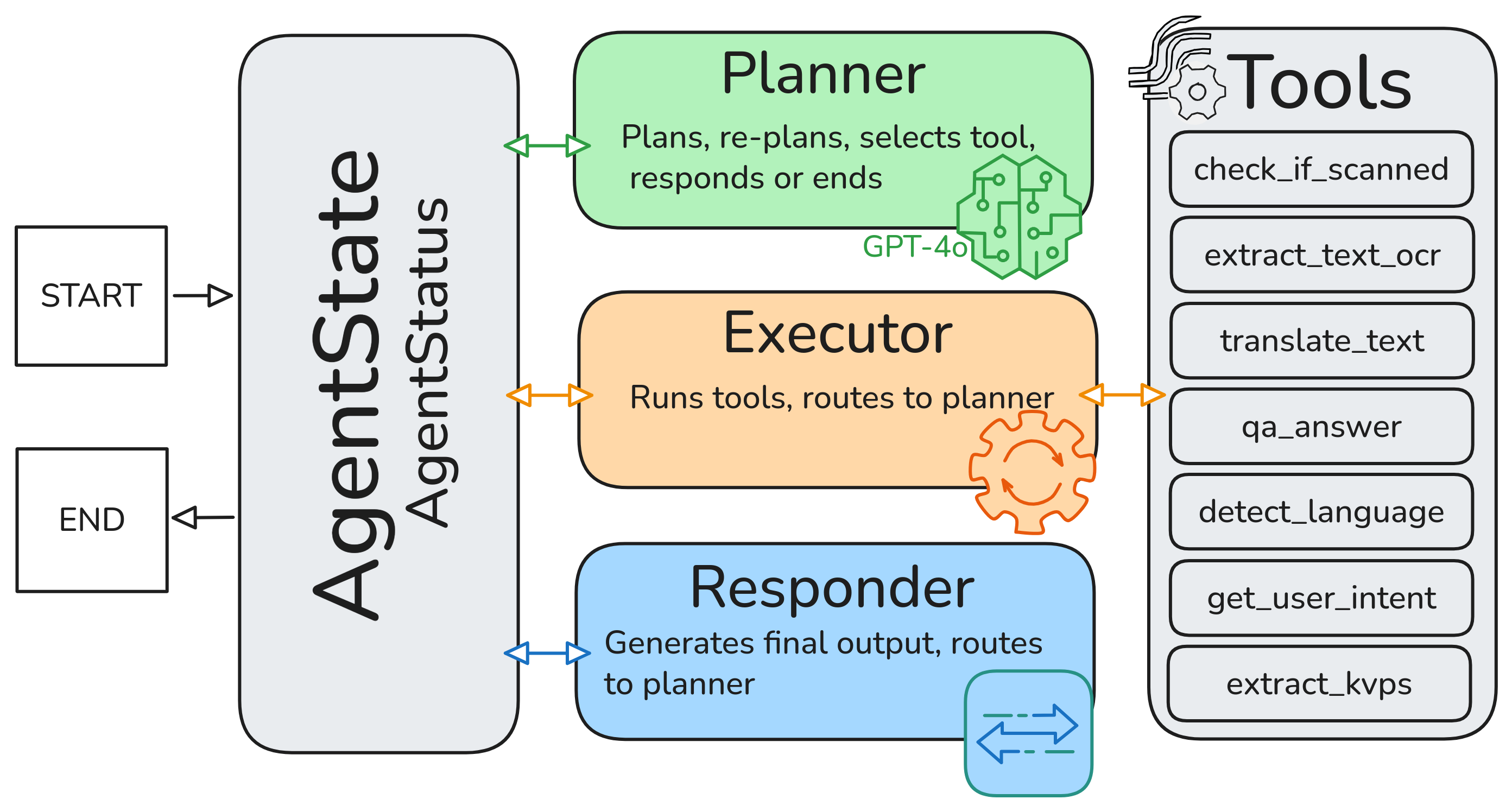}
    \caption{High-level architecture with planner, executor, responder, and control flow via AgentState.
    }
    \label{fig:system_overview}
    \rmspace
\end{figure}

Our system supports multilingual KIE extraction and QA over diverse DoP document formats. The architecture is built around three core components; \textbf{planner}, \textbf{executor}, and \textbf{responder}, coordinated through a shared memory model (\texttt{AgentState}) and a control flag (\texttt{AgentStatus}). These components work together to enable dynamic, context-aware reasoning and robust tool orchestration. A high-level overview of the system is shown in Figure~\ref{fig:system_overview}.

\subsection{AgentState and Execution Flow}
The system uses a shared memory object \texttt{AgentState} to track all relevant information during execution. It stores user input, document metadata, tool outputs, and reasoning history, allowing the planner, executor, and responder to make informed, context-aware decisions.

Transitions between stages are managed by \texttt{AgentStatus}, which can take one of five values: \texttt{PLAN} (trigger planner reasoning), \texttt{NEED_TOOL} (trigger executor to run the selected tool), \texttt{RESPOND} (trigger responder to generate final output), \texttt{SUCCESS} (task completed with success), or \texttt{END} (workflow terminated without success).

The system is implemented as a \textit{directed state graph}, where each node corresponds to a specific function. All nodes read and update the AgentState. The planner inspects it to decide the next step, the executor runs tools and integrates results, and the responder produces the final output or defers control back to planning.
This state-driven control loop enables robust, adaptive reasoning: each component operates with full visibility of the task context, supports fallback and recovery strategies, and enables adaptive execution paths.

\subsection{Planning Module}
The planner selects the next action based on the current AgentState. At each reasoning step, it evaluates the accumulated information and decides whether to invoke a tool, generate a response, or terminate the workflow.

\begin{figure}[ht]
   \centering
   \includegraphics[width=\columnwidth]{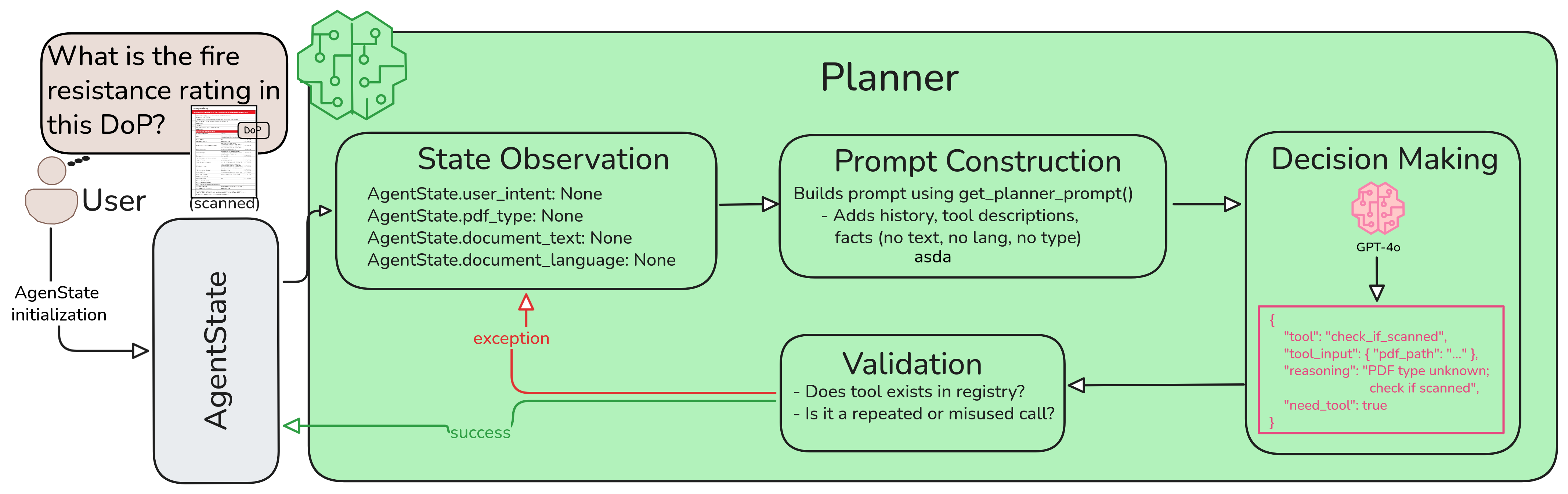}
\caption{Planner workflow during a single reasoning cycle, triggered when \texttt{AgentStatus} is set to \texttt{PLAN}}
   \label{fig:planner-detailed}
   \rmspace
\end{figure}

The next action selection decision is made by prompting an LLM with a structured input that includes the current state, tool metadata, and relevant facts. The model returns a structured JSON object specifying the planner’s reasoning, the next tool (if any), and its input parameters (Figure~\ref{fig:planner-detailed}).

To ensure reliability, the planner enforces several safeguards: (1) it checks for required values (e.g., \texttt{document\_language}) before allowing certain tools; (2) it halts repeated tool use with identical inputs to prevent loops; (3) it detects misuse, such as selecting a QA tool for verification; and (4) it performs recovery by incorporating error context from failed tool calls.

Once a decision is made, the planner updates \texttt{AgentStatus} to reflect the next step: \texttt{NEED\_TOOL} if a tool should be executed, \texttt{RESPOND} if the task is complete, \texttt{PLAN} if the decision process fails or needs to be retried, and \texttt{END} if an unrecoverable error, such as loop detection or repeated misuse, occurs.

\subsection{Executor Module}

\begin{figure}[ht]
   \centering
   \includegraphics[width=\columnwidth]{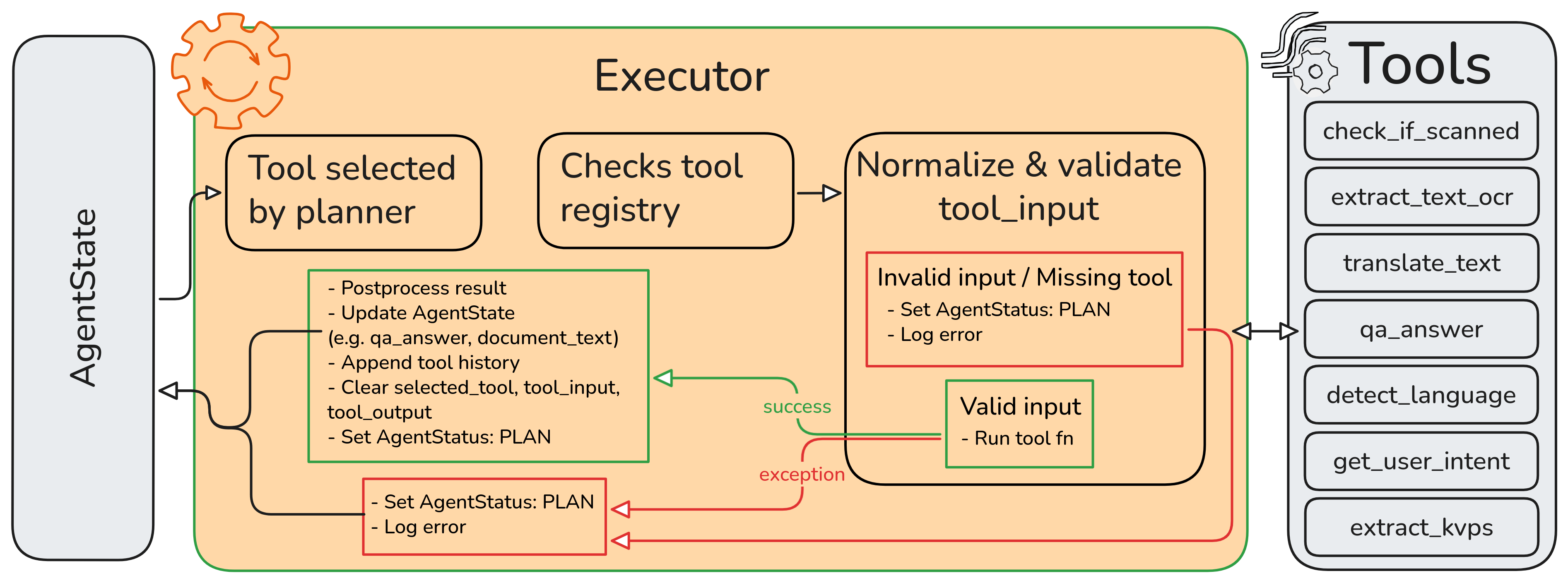}
\caption{Executor workflow during a single reasoning cycle, triggered when \texttt{AgentStatus} is set to \texttt{NEED_TOOL}.}
   \label{fig:executor-detailed}
   \rmspace
\end{figure}

The executor translates the planner’s high-level decisions into concrete actions by running the selected tool and integrating its results into the AgentState. Once a tool and its input are specified, the executor retrieves the corresponding tool and normalizes the input using the tool’s schema to ensure compatibility (Figure~\ref{fig:executor-detailed}).

Tools are executed via a uniform interface, allowing flexibility in implementation. On success, tool outputs are postprocessed and written to relevant fields in the AgentState (e.g., \texttt{user_language}, \texttt{document_text}, \texttt{extracted_kvps}, \texttt{qa_answer}, \texttt{verification_result}). On failure-due to parsing errors, misuse, runtime issues, or external constraints-the executor updates the AgentState with error metadata and stores the faulty output. In both cases, \texttt{AgentStatus} is set to \texttt{PLAN}, signaling that a new reasoning cycle should begin.
To prevent possible infinite loops, the executor tracks the last executed tool and its output, enabling the planner to adjust its strategy.

\subsection{Responder Module}
The responder serves as a quality gate, returning outputs only when verified.
The verification status is set by an LLM-based verification step that receives both the original document and the extracted output, performing semantic validation to confirm the factual grounding. Unlike rule-based validation, this approach handles paraphrasing, OCR artifacts, and multi-section synthesis.
Upon successful verification, the responder returns \texttt{translated\_output} when available (matching user language), or the raw extraction.
If verification fails, the responder withholds results and returns control to the planner, triggering adaptive replanning. This verification-driven feedback loop ensures only complete, high-confidence outputs reach the user, critical for sensitive applications such as construction compliance.

\begin{figure}[ht]
   \centering
   \includegraphics[width=\columnwidth]{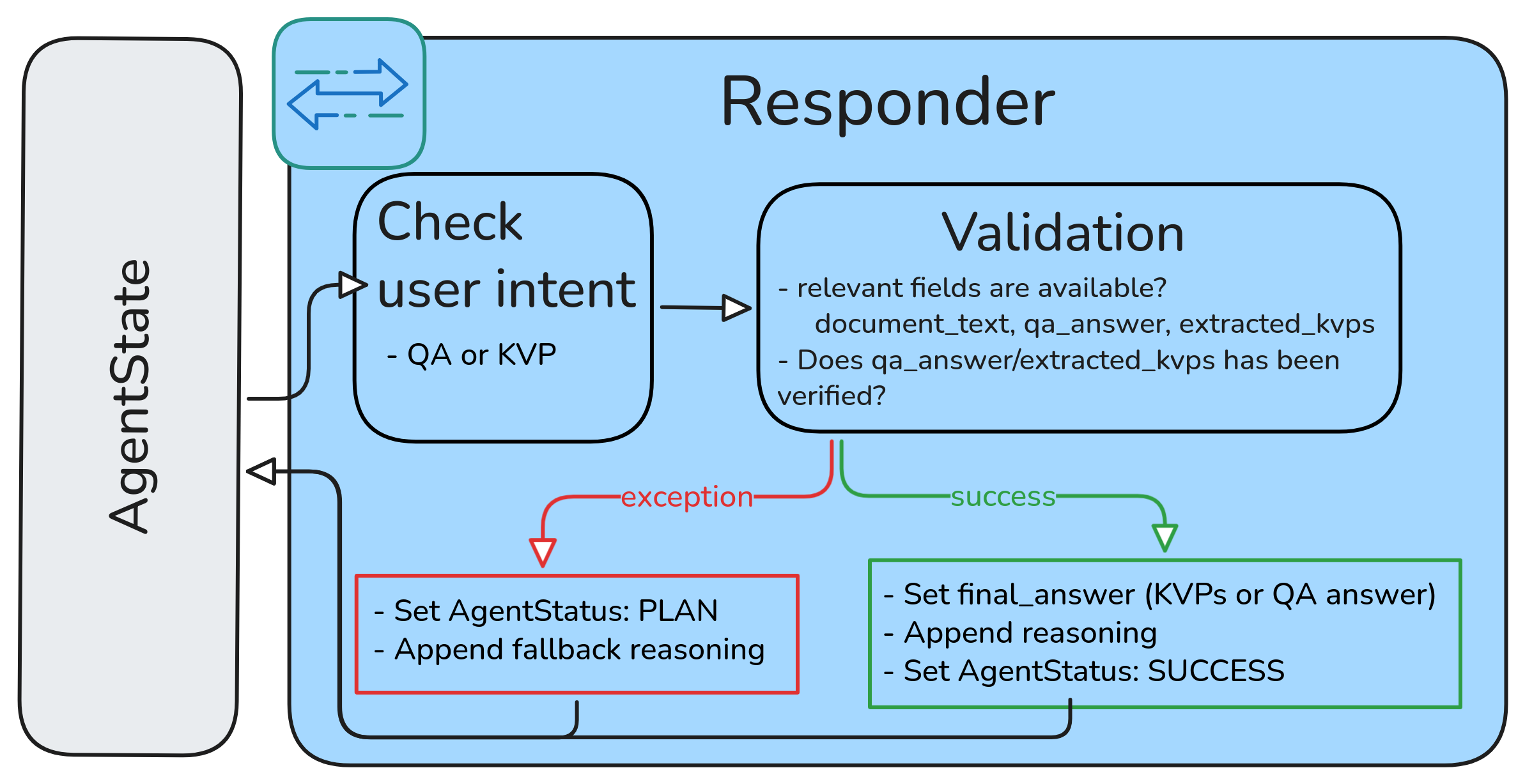}
\caption{Responder workflow during a single reasoning cycle, triggered when \texttt{AgentStatus} is set to \texttt{RESPOND}.}
   \label{fig:responder-detailed}
   \rmspace
\end{figure}

\section{Experimental Setup}
\label{sec:experimental_setup}

Our evaluation is structured around two primary dimensions: (i) \textit{Schema Adherence (SA)}: the validity of the extracted entity types; and (ii) \textit{EM}: the correctness of the predicted values against the ground truth. Unlike other evaluation strategies that focus on intermediate planning or memory steps~\cite{yehudai2025surveyevaluationllmbasedagents}, we strictly assess the final output fidelity against manually-annotated ground truth.

\paragraph{Dimension 1: Schema Adherence (SA).}
To evaluate whether the agent extracts the correct entity types (irrespective of the values), we define the \textit{SA score}. Let \( E_{\text{gt}} \) be the set of ground-truth entity types for a given document \( d \), and \( E_{\text{pred}} \) be the set of entity types predicted by the agent:

\begin{equation}
\small
\text{SA} = 
\frac{|E_{\text{gt}} \cap E_{\text{pred}}|}
{|E_{\text{gt}}|}
\label{eq:schema_adherence}
\end{equation}

This metric acts as a strict recall measure for the document schema, penalizing hallucinated or missing entity types.

\paragraph{Dimension 2: Exact Match (EM).}
To evaluate the correctness of the extracted values, we compute the \textit{EM score} over the ground-truth entity types. For each entity type \( e \in E_{\text{gt}} \), let \( v_e \) be the ground-truth value and \( \hat{v}_e \) be the predicted value:

\begin{equation}
\small
\text{EM} = \frac{1}{|E_{\text{gt}}|} \sum_{e \in E_{\text{gt}}} \mathbb{I}(\text{norm}(\hat{v}_e) = \text{norm}(v_e))
\label{eq:exact_match}
\end{equation}

where \( \mathbb{I}(\cdot) \) is the indicator function, and \( \text{norm}(\cdot) \) represents standard string normalization (trimming whitespace and lowercasing). If the agent fails to extract an entity type \( e \) (i.e., \( e \notin E_{\text{pred}} \)), we assume \( \hat{v}_e = \varnothing \), resulting in a score of 0 for that entity.

\subsection{Baseline Methods}
\label{sec:baseline_settings}
To evaluate our AgenticIE approach, we compare it against zero-shot baselines using GPT-4o (text and vision). Our methods are designed based on best-practice configurations (i.e., input preparation and prompt design) from a recent evaluation study of IE from layout-rich documents~\cite{colakoglu-etal-2025-problem}. We do not include traditional Visually Rich Document Understanding (VRDU) models such as LayoutLMv3 as they struggle fundamentally with generative/open-ended extraction where the keys are not fixed in advance. This makes them architecturally unsuitable for open schema settings.

\textbf{Input Preparation}
For each document, we extract text using a hybrid pipeline that first attempts native PDF text extraction and falls back to OCR when necessary. A lightweight heuristic detects garbage output (e.g., non-printable characters), prompting page-level OCR recovery. For vision-based baselines, documents are rendered into high-resolution images (300 DPI) and encoded as Base64 PNGs to comply with GPT-4o-Vision input requirements. All inputs are passed directly to the LLMs along with handcrafted prompts.

\textbf{Prompt Design}
For both the KIE and QA, we evaluate the baseline models in English and German across two modalities: text-only and vision (using document images).
For KIE, we use a constrained prompt that includes a predefined schema with the entity types to guide the model's extraction. For the QA task, a unified prompt asks the model to extract a concise answer, return a list if needed, or an empty string if no information is found.
Across all tasks, models are instructed to respond in the same language as the prompt and to return valid JSON for evaluation. For detailed prompt examples, please see Appendix~\ref{app:baseline_prompts} and~\ref{appendix:qa-baseline-vision-en}.

\section{Experimental Evaluation}
\label{sec:evaluation}

We evaluate our system in comparison to the baseline methods separately for KIE (Section~\ref{sub:kie-evaluation}) and QA (Section~\ref{sub:qa-evaluation}). Last, we investigate runtime, cost and token usage (Section~\ref{sub:runtime-evaluation}).

\subsection{Results for the KIE Task}
\label{sub:kie-evaluation}

We evaluate extraction performance on the 12 primary keys defined in Section~\ref{sec:target_fields}. Consistent with our \textbf{Hybrid Schema} definition, we report results on two distinct subsets of keys: \textbf{Fixed Schema:} The set of standard metadata fields (e.g., \textit{Manufacturer}, \textit{Declaration Number}) where the schema is static and known a priori.
\textbf{Open Schema:} The complex, nested fields (e.g., \textit{Declared Performance}) where the agent must dynamically discover the structure and sub-keys without a predefined template.

\subsubsection{Schema Adherence.}
This metric evaluates structural correctness by measuring the intersection between predicted and ground-truth entity types (Table~\ref{tab:schema_adherence_results} ). 

In the \textbf{Fixed Schema} setting, the Agent achieves perfect reliability (1.0) across all language pairs, effectively solving the metadata extraction task. The baselines are also strong, though GPT-4o-V shows slight degradation (0.941) on German documents, likely due to OCR challenges in the vision encoder.

The \textbf{Open Schema} setting reveals the core advantage of the agentic approach. Discovery of nested keys in \textit{Declared Performance} is significantly harder, particularly in cross-lingual scenarios. While baselines fluctuate wildly, e.g., GPT-4o drops to 0.470 on En $\to$ De, the Agent maintains robust performance (0.633), achieving the highest average adherence (0.613). This confirms that the planner-executor loop effectively adapts to dynamic schemas and mixed-language scenarios, where static prompting fails.

\begin{table}[hbt]
\centering
\footnotesize
\caption{SA. The \textbf{Agent} achieves perfect adherence in fixed settings and significantly outperforms baselines in cross-lingual open-schema extraction ($Q$: Query Language, $D$: Document Language).}
\label{tab:schema_adherence_results}
\resizebox{\columnwidth}{!}{%
\begin{tabular}{llccc}
\toprule
\textbf{Task} & \textbf{$\mathbf{Q} \to \mathbf{D}$} & \textbf{GPT-4o} & \textbf{GPT-4o-V} & \textbf{Agent (Ours)} \\
\midrule
\multirow{5}{*}{\textbf{\shortstack{Fixed\\Schema}}} 
 & En $\to$ En & \textbf{1.000} & \textbf{1.000} & \textbf{1.000} \\
 & En $\to$ De & \textbf{1.000} & 0.961 & \textbf{1.000} \\
 & De $\to$ En & 0.964 & \textbf{1.000} & \textbf{1.000} \\
 & De $\to$ De & \textbf{1.000} & 0.941 & \textbf{1.000} \\
 \cmidrule(lr){2-5}
 & \textit{Mean} & \textit{0.991} & \textit{0.976} & \textbf{\textit{1.000}} \\
\midrule
\multirow{5}{*}{\textbf{\shortstack{Open\\Schema}}} 
 & En $\to$ En & 0.596 & \textbf{0.640} & 0.592 \\
 & En $\to$ De & 0.470 & 0.386 & \textbf{0.633} \\
 & De $\to$ En & 0.485 & 0.498 & \textbf{0.571} \\
 & De $\to$ De & \textbf{0.677} & 0.526 & 0.654 \\
 \cmidrule(lr){2-5}
 & \textit{Mean} & \textit{0.557} & \textit{0.513} & \textbf{\textit{0.613}} \\
\bottomrule
\end{tabular}
}
\rmspace
\end{table}

\subsubsection{Exact Match (EM)}

We evaluate value correctness using the EM metric. As shown in Table~\ref{tab:kie_results}, the Agent consistently outperforms baselines, particularly in cross-lingual tasks.

For \textbf{Fixed Schema} keys, the Agent achieves a mean EM of 0.411, surpassing GPT-4o (0.333) and GPT-4o-V (0.327). The gap is most visible in the monolingual German setting (0.583 vs. 0.430), suggesting that the agent's tool use (specifically tailored OCR and translation tools) yields cleaner value extraction than end-to-end generation.

In the \textbf{Open Schema} task, precision is critical as incorrect nesting invalidates the extracted value. The Agent achieves the highest mean EM (0.321). Notably, in the difficult En $\to$ De setting (English query, German document), the Agent scores 0.352, more than doubling the performance of GPT-4o (0.143). This demonstrates that our agentic architecture successfully bridges the language gap for complex, nested data structures where baselines struggle to align the user's English intent with the document's German hierarchy.

\begin{table}[hbt]
\centering
\footnotesize
\caption{KIE EM. The \textbf{Agent} demonstrates superior cross-lingual stability, particularly in English $\to$ German tasks. ($Q$: Query Language, $D$: Document Language).}
\label{tab:kie_results}
\resizebox{\columnwidth}{!}{%
\begin{tabular}{llccc}
\toprule
\textbf{Task} & $\mathbf{Q} \to \mathbf{D}$ & \textbf{GPT-4o} & \textbf{GPT-4o-V} & \textbf{Agent (Ours)} \\
\midrule
\multirow{5}{*}{\textbf{\shortstack{Fixed\\Schema}}} 
 & En $\to$ En & 0.351 & 0.396 & \textbf{0.423} \\
 & En $\to$ De & 0.242 & 0.219 & \textbf{0.292} \\
 & De $\to$ En & 0.310 & 0.307 & \textbf{0.346} \\
 & De $\to$ De & 0.430 & 0.387 & \textbf{0.583} \\
 \cmidrule(lr){2-5}
 & \textit{Mean} & \textit{0.333} & \textit{0.327} & \textbf{\textit{0.411}} \\
\midrule
\multirow{5}{*}{\textbf{\shortstack{Open\\Schema}}} 
 & En $\to$ En & 0.258 & 0.310 & \textbf{0.320} \\
 & En $\to$ De & 0.143 & 0.136 & \textbf{0.352} \\
 & De $\to$ En & 0.189 & 0.224 & \textbf{0.244} \\
 & De $\to$ De & \textbf{0.371} & 0.298 & 0.367 \\
 \cmidrule(lr){2-5}
 & \textit{Mean} & \textit{0.240} & \textit{0.242} & \textbf{\textit{0.321}} \\
\bottomrule
\end{tabular}
}
\rmspace
\end{table}

\subsection{Results for the QA Task}
\label{sub:qa-evaluation}

We evaluate QA capabilities using the same EM metric, focusing on the ability to retrieve precise answers for both static metadata (Fixed Schema) and dynamic properties (Open Schema) (Table~\ref{tab:qa_results}).

\begin{table}[hbt]
\centering
\footnotesize
\caption{QA EM. The \textbf{Agent} outperforms baselines in both Fixed and Open schema settings, showing particular benefit in German-language tasks. ($Q$: Query Language, $D$: Document Language).}
\label{tab:qa_results}
\resizebox{\columnwidth}{!}{%
\begin{tabular}{llccc}
\toprule
\textbf{Task} & $\mathbf{Q} \to \mathbf{D}$ & \textbf{GPT-4o} & \textbf{GPT-4o-V} & \textbf{Agent (Ours)} \\
\midrule
\multirow{5}{*}{\textbf{\shortstack{Fixed\\Schema}}} 
 & En $\to$ En & 0.503 & 0.500 & \textbf{0.570} \\
 & En $\to$ De & \textbf{0.393} & 0.379 & 0.329 \\
 & De $\to$ En & 0.430 & 0.430 & \textbf{0.431} \\
 & De $\to$ De & 0.444 & 0.453 & \textbf{0.515} \\
 \cmidrule(lr){2-5}
 & \textit{Mean} & \textit{0.443} & \textit{0.441} & \textbf{\textit{0.461}} \\
\midrule
\multirow{5}{*}{\textbf{\shortstack{Open\\Schema}}} 
 & En $\to$ En & 0.367 & 0.262 & \textbf{0.384} \\
 & En $\to$ De & 0.292 & 0.212 & \textbf{0.384} \\
 & De $\to$ En & \textbf{0.307} & 0.211 & \textbf{0.307} \\
 & De $\to$ De & 0.446 & 0.308 & \textbf{0.479} \\
 \cmidrule(lr){2-5}
 & \textit{Mean} & \textit{0.353} & \textit{0.248} & \textbf{\textit{0.389}} \\
\bottomrule
\end{tabular}
}
\rmspace
\end{table}

\textbf{Fixed Schema QA.}
Here, language alignment is a strong predictor of performance. For matched languages ($En \to En$ and $De \to De$), the Agent leads significantly, achieving 0.570 and 0.515 EM scores respectively. This suggests that its explicit verification step filters out hallucinations more effectively than the baselines. A notable exception is the $En \to De$ setting (English query on German document), where GPT-4o (0.393) outperforms the Agent (0.329). This dip likely stems from the ``translation-then-retrieval'' bottleneck: minor errors in translating a specific technical query (e.g., "Notified Body") into German can cause retrieval failures that end-to-end models might bypass through soft alignment.

\textbf{Open Schema QA.}
The Open Schema task targets values deeply nested within complex tables (e.g., \textit{Declared Performance}). Here, the Agent's architectural benefits are most pronounced. It achieves the highest mean EM score (0.389), outperforming GPT-4o (0.353) and GPT-4o-V (0.248).
The results highlight a consistent \textbf{cross-lingual asymmetry}: extracting answers from German documents is systematically harder when the query is in English ($En \to De$). However, the Agent mitigates this better in the open schema setting (0.384) compared to GPT-4o (0.292) and GPT-4o-V (0.212). The vision baseline struggles significantly here, likely because visual encoders often fail to parse the dense, multi-page tables typical of German DoPs.

\subsection{Runtime, Cost and Token Comparison}
\label{sub:runtime-evaluation}

\begin{figure}[ht]
    \centering
\includegraphics[width=1.0\linewidth]{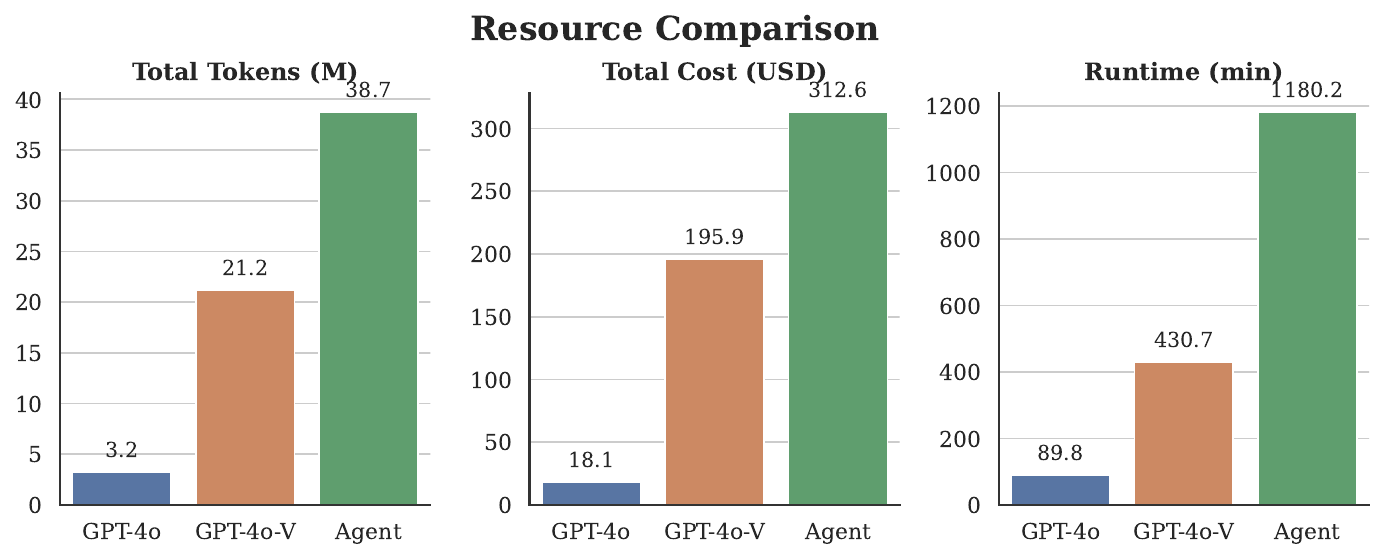}
\caption{Resource usage. The agentic system incurs significantly higher costs compared to static baselines.}
\label{fig:resource_comparison}
\rmspace
\end{figure}

Last, we compare the computational efficiency of the agentic system against both baseline models in terms of token usage, total cost, and runtime. Figure~\ref{fig:resource_comparison} %
summarizes these metrics across the KIE and QA tasks. Input and output token counts are derived from the prompt and completion segments. The cost is computed based on the GPT-4o pricing (September 2025), and the run time reflects the total wall clock time per tasks. Although AgenticIE provides performance benefits (see Section~\ref{sub:kie-evaluation} and \ref{sub:qa-evaluation}), it has the highest resource consumption. This is mainly due to the iterative nature of the agentic loop, where the planner and executor repeatedly exchange state information and tool outputs.
High resource use of agentic AI is a known and open research problem~\cite{nooralahzadeh2024explainable}. Future work must focus on optimizing the planning horizon and reducing redundant tool calls to make agentic IE more sustainable for large-scale IE applications.

\section{Related Work}
\label{sec:related_work}

\noindent \textbf{Visually Rich Document Understanding.}
Layout-aware models~\cite{xu2020layoutlm, xu2021layoutxlm} and multimodal datasets~\cite{jaume2019funsd, park2019cord, huang2019icdar2019} combine text, layout, and visual features for structured extraction. However, many datasets and methods consist of clean, monolingual, and semantically regular documents. In contrast, \textit{DoP documents exhibit high variability in layout, quality, and language}.

\noindent \textbf{Prompt-Based Extraction.}
Prompt-based methods frame extraction as question answering~\cite{hu2023question}, incorporate layout structure~\cite{lin2023layoutprompter}, and directly decode fields from images~\cite{kim2021donut}. These methods reduce reliance on fixed schemas, but often assume consistently formatted, high-quality input. Further, they require a \textit{customization of the extraction pipeline to the specific task and dataset}~\cite{colakoglu-etal-2025-problem}.

\noindent \textbf{Tool-Augmented Reasoning and Modular Agents.}
Recent studies enhance LLMs with tool use~\cite{schick2023toolformer, wang2024agent}, but most document pipelines remain static, relying on fixed OCR and parsing sequences~\cite{luo2024layoutllm, perot2023lmdx}. More flexible designs~\cite{zhang2024knowvrdu, neupane-etal-2025-structured} introduce partial modularity (e.g., knowledge guidance or layout-aware prompts), whereas they lack runtime planning or tool orchestration. Agentic systems~\cite{patil2023gorillalargelanguagemodel, qin2023toolllmfacilitatinglargelanguage, gao2023palprogramaidedlanguagemodels} decouple planning, tool use, and response generation. \textit{Benchmarks such as AgentBench~\cite{liu2023agentbench} evaluate agents in interactive environments, but not for IE.}

\noindent \textbf{Planning, Feedback, and Memory in Agents.}
Recent patterns include task decomposition, plan revision, and tool routing~\cite{huang2024understanding, masterman2024landscapeemergingaiagent}. Iterative refinement methods implement feedback loops for recovery~\cite{shinn2023reflexion, madaan2023self, liu2023llm+}, which \textit{our agent mirrors via a planning-execution-feedback cycle to handle OCR or parsing failures}.
Structured planning frameworks support modular reasoning and collaborative workflows~\cite{wang2023describe, wu2024autogen}, while code-trained LLMs offer stronger schema-aware planning~\cite{yang2024if}, motivating our choice of a structured planner. \textit{Long-term memory mechanisms}, such as scratchpads or persistent state~\cite{zhang2024survey}, \textit{are integrated in our \texttt{AgentState} to track modality, task intent, and tool outcomes. Retrieval-augmented planners~\cite{ruan2023tptulargelanguagemodelbased, kong2023tptu} also inform our approach to runtime tool selection across diverse document types.}

\section{Conclusion}
\label{conclusion}
This paper introduces AgenticIE, as the first agentic system and extensive dataset for IE from DoP documents, with a focus on handling hybrid schemas in English and German. Our agent outperforms static and multimodal LLM baselines, achieving higher EM scores compared to GPT-4o and GPT-4o-V across KIE and QA. The experimental analysis showcase the challenging nature of NLP applications to DoP.

\clearpage
\section*{Limitations}
\label{sec:limitations}

This work focuses on adaptive information extraction within a specific regulatory context-Declaration of Performance (DoP) documents-characterized by multilinguality and diverse layouts. The agentic system is therefore intentionally designed for domain-grounded reasoning rather than general-purpose intelligence. Extending it to other domains such as financial or legal records would require retraining the planner and adapting the tool registry to new schemas and reasoning conventions.

Our evaluation emphasizes structural and semantic correctness but does not address higher-order reasoning such as factual consistency, regulatory validation, or human-in-the-loop correction. Exploring such capabilities would shift the system from extraction toward auditable compliance reasoning. Likewise, while the agent performs reliably across English and German, scaling to low-resource or morphologically complex languages would demand cross-lingual schema alignment and translation-aware planning.

Finally, the current design focuses on textual content and inferred structure. Incorporating explicit visual cues such as layout topology or table geometry could further enhance precision in complex or scanned documents, but doing so would require multimodal model integration beyond this paper’s technical scope.

\section*{Ethical Considerations}
The development and deployment of this agentic system for extracting information from Declaration of Performance (DoP) documents require careful consideration of several ethical dimensions, including data privacy, potential for misuse, algorithmic bias, and environmental impact.

\paragraph{Data Privacy and Personal Information.}
As noted, DoP documents are public records mandated by EU regulation, but they can contain personal identifiers, such as the names, positions, and signatures of the individuals responsible for the declaration. While this information is publicly accessible, its automated collection and processing at scale warrant ethical scrutiny. Our approach mitigates privacy risks by using this data strictly for research purposes to evaluate the extraction system.

\paragraph{Reliability and Risk of Misuse}
The information extracted from DoPs, such as fire resistance ratings or the use of hazardous substances, is critical for ensuring the safety and compliance of construction projects. An error in extraction could lead to the selection of non-compliant materials, posing significant safety risks. Therefore, this system is intended to be an assistive tool for engineers and architects, not a fully autonomous replacement for human oversight. End-users should be made aware that the system's output must be verified by a qualified professional before being used for any critical decisions. The system's built-in verification steps and high performance are designed to minimize errors, but the ``human-in-the-loop'' remains essential for accountability in high-stakes applications.

\paragraph{Algorithmic Bias and Fairness}
Our dataset comprises documents in English and German. While the agentic system demonstrates more stable cross-lingual performance than the baselines, the paper notes that performance can still vary between languages and in cross-lingual scenarios. Deploying such a system across the multilingual European market could lead to performance disparities, potentially disadvantaging users working with less-represented languages. The data was also collected via web search and from specific repositories, which may introduce a selection bias towards certain manufacturers or product types. Future work should aim to expand the dataset to include a wider range of languages and document origins to ensure more equitable and robust performance.

\paragraph{Computational Cost and Environmental Impact}
The experiments show that the proposed agentic system, while more accurate, incurs a significantly higher computational cost in terms of token usage, cost, and runtime compared to baseline models. The total token count for the agent was 38.69 million, costing over \$300 and taking nearly 20 hours to run. We acknowledge that the energy consumption associated with such extensive computation has an environmental impact. This trade-off between performance and efficiency is a known challenge for agentic architectures. We advocate for continued research into more computationally efficient models and methods to create more sustainable solutions for information extraction.

\paragraph{AI Assistants In Research Or Writing}
We have used GPT-5, Gemini 2.5 pro and Claude Opus 4.1 to help us refine our writing. All final text in the paper has been verified by the authors.

\bibliography{bibliography.bib}

\clearpage
\appendix %
\section{Appendix}

This appendix provides supplementary materials to support the reproducibility and deeper understanding of our AgenticIE system and the DoP dataset. We first detail the dataset construction process, including the target annotation keys, QA generation templates, and concrete examples of Key Information Extraction (KIE) and Question Answering (QA) pairs. Subsequently, we provide implementation specifics for the agentic architecture, covering the tool registry, prompt design strategies, and schema definitions. Finally, we present the full prompts used for the baseline models and a qualitative error analysis of extraction failures.

\subsection{Dataset Annotation Details}
\subsubsection{Target Keys}
\label{appendix:12-target-keys}

The following JSON-style structure presents list of 12 target keys selected for annotation and extraction from DoP documents, including their English and German versions. These keys represent core elements required for regulatory compliance.

\begin{tcolorbox}[
    breakable,
    minted language=python,
    colback=orange!5,
      colframe=brown!100,
      boxrule=0.4pt,
    title=12 Target Keys in EN-DE
]
\begin{lstlisting}[language=json, breaklines=true, basicstyle=\ttfamily\footnotesize]
[
  { "key_en": "Declaration Number", "key_de": "Erklärungsnummer" },
  { "key_en": "Unique Identification Code of the Product-Type", "key_de": "Eindeutiger Identifikationscode des Produkttyps" },
  { "key_en": "Intended Use(s)", "key_de": "Vorgesehene Verwendung" },
  { "key_en": "Manufacturer", "key_de": "Hersteller" },
  { "key_en": "Authorised Representative", "key_de": "Bevollmächtigter" },
  { "key_en": "System(s) of Assessment and Verification of Constancy of Performance (AVCP)", "key_de": "System(e) zur Bewertung und Überprüfung der Leistungsbeständigkeit (AVCP)" },
  { "key_en": "Harmonised Standard", "key_de": "Harmonisierte Norm" },
  { "key_en": "Notified Body", "key_de": "Notifizierte Stelle" },
  { "key_en": "Declared Performance", "key_de": "Erklärte Leistung" },
  { "key_en": "Appropriate Technical Documentation and/or Specific Technical Documentation", "key_de": "Geeignete technische Dokumentation und/oder besondere technische Dokumentation" },
  { "key_en": "Declaration Statement", "key_de": "Erklärungstext" },
  { "key_en": "Signature", "key_de": "Unterschrift" }
]
\end{lstlisting}
\end{tcolorbox}

\subsubsection{KIE Annotation Sample}
\label{appendix:kie-annotation}
Example of a multilingual KIE annotation for the key "Declared Performance" from a German-language DoP document.
\begin{tcolorbox}[
    breakable,
    minted language=python,
    colback=orange!5,
      colframe=brown!100,
      boxrule=0.4pt,
    title=KIE annotation for the key "Declared Performance"
]
\begin{lstlisting}[language=json, breaklines=true, basicstyle=\ttfamily\footnotesize]
{
  "doc_name": "URSA DF 39.pdf",
  "key_en": "Declared Performance",
  "key_de": "Erklaerte Leistung",
  "value_en": {
    "Thermal Resistance (RD)": "1.00 - 6.15 m^2K/W (for nominal thickness dN 40-240 mm, T2)",
    "Thermal Conductivity (lambda_D)": "0.039 W/mK",
    "Reaction to Fire": "Class A1 (for thickness 40-240 mm)",
    "Durability of Thermal Resistance": "Stable under heat, weathering, and aging",
    "Durability of Reaction to Fire": "Stable over time, not affected by organic content",
    "Dimensional Stability": "DS (70,-) <=1%
    "Water Vapor Permeability": "MU1",
    "Airborne Sound Insulation": "Airflow resistance AFr5 >= 5 kPa s/m^2",
    "Smoldering Behavior": "NPD, European test under development",
    "Release of Dangerous Substances": "NPD, European test under development"
  },
  "value_de": {
    "Waermedurchlasswiderstand (RD)": "1.00 - 6.15 m^2K/W (fuer Dicken dN 40-240 mm, T2)",
    "Waermeleitfaehigkeit (lambda_D)": "0.039 W/mK",
    "Brandverhalten": "Klasse A1 (fuer Dicke 40-240 mm)",
    "Dauerhaftigkeit des Waermedurchlasswiderstandes": "Stabil gegen Hitze, Witterung und Alterung",
    "Dauerhaftigkeit des Brandverhaltens": "Stabil im Laufe der Zeit, nicht durch organischen Gehalt beeinflusst",
    "Dimensionsstabilitaet": "DS (70,-) <=1%
    "Wasserdampfdurchlaessigkeit": "MU1",
    "Luftschalldaemmung": "Stroemungswiderstand AFr5 >= 5 kPa s/m^2",
    "Glimmverhalten": "NPD, Europaeisches Pruefverfahren in Entwicklung",
    "Freisetzung gefaehrlicher Stoffe": "NPD, Europaeisches Pruefverfahren in Entwicklung"
  },
  "doc_lang": "de"
}
\end{lstlisting}
\end{tcolorbox}

The following JSON-style structure presents extracted key-value pairs (KIEs) from the document \textit{Sager\_ISO-SWISS-Design-Panels.pdf}, annotated in both English and German for evaluation and training purposes.

\begin{tcolorbox}[
    breakable,
    minted language=python,
    colback=orange!5,
      colframe=brown!100,
      boxrule=0.4pt,
    title=KIE Annotation Example
]
\begin{lstlisting}[language=json, breaklines=true, basicstyle=\ttfamily\footnotesize]
[
  {
    "doc_name": "Sager_ISO-SWISS-Design-Panels.pdf",
    "key_en": "Declaration Number",
    "key_de": "Erklärungsnummer",
    "value_en": "sa-0008-isoswiss-pdk-a1-210228",
    "value_de": "sa-0008-isoswiss-pdk-a1-210229",
    "doc_lang": "de"
  },
  {
    "doc_name": "Sager_ISO-SWISS-Design-Panels.pdf",
    "key_en": "Unique Identification Code of the Product-Type",
    "key_de": "Eindeutiger Identifikationscode des Produkttyps",
    "value_en": "sa-0008-isoswiss-pdk-a1-210228",
    "value_de": "sa-0008-isoswiss-pdk-a1-210229",
    "doc_lang": "de"
  },
  {
    "doc_name": "Sager_ISO-SWISS-Design-Panels.pdf",
    "key_en": "Intended Use(s)",
    "key_de": "Vorgesehene Verwendung",
    "value_en": "Thermal insulation for buildings",
    "value_de": "Wärmedämmstoffe für Gebäude",
    "doc_lang": "de"
  },
  {
    "doc_name": "Sager_ISO-SWISS-Design-Panels.pdf",
    "key_en": "Manufacturer",
    "key_de": "Hersteller",
    "value_en": "Sager AG, Dornhügelstrasse 10, CH-5724 Dürrenäsch, Switzerland",
    "value_de": "Sager AG, Dornhügelstrasse 10, CH-5724 Dürrenäsch, Schweiz",
    "doc_lang": "de"
  },
  {
    "doc_name": "Sager_ISO-SWISS-Design-Panels.pdf",
    "key_en": "Authorised Representative",
    "key_de": "Bevollmächtigter",
    "value_en": "Not applicable",
    "value_de": "Nicht angegeben",
    "doc_lang": "de"
  },
  {
    "doc_name": "Sager_ISO-SWISS-Design-Panels.pdf",
    "key_en": "System(s) of Assessment and Verification of Constancy of Performance (AVCP)",
    "key_de": "System(e) zur Bewertung und Überprüfung der Leistungsbeständigkeit (AVCP)",
    "value_en": "System 3; System 1",
    "value_de": "System 3; System 2",
    "doc_lang": "de"
  },
  {
    "doc_name": "Sager_ISO-SWISS-Design-Panels.pdf",
    "key_en": "Harmonised Standard",
    "key_de": "Harmonisierte Norm",
    "value_en": "EN 13162:2012 + A1:2015",
    "value_de": "EN 13162:2012 + A1:2016",
    "doc_lang": "de"
  },
  {
    "doc_name": "Sager_ISO-SWISS-Design-Panels.pdf",
    "key_en": "Notified Body",
    "key_de": "Notifizierte Stelle",
    "value_en": "FIW München (Kennnummer 0751)",
    "value_de": "FIW München (Kennnummer 0751)",
    "doc_lang": "de"
  },
  {
    "doc_name": "Sager_ISO-SWISS-Design-Panels.pdf",
    "key_en": "Declared Performance",
    "key_de": "Erklärte Leistung",
    "value_en": {
      "Thermal Resistance (RD)": "1.25 - 6.45 m2K/W (for nominal thickness dN 40-200 mm, T5)",
      "Thermal Conductivity (<=lambdaD)": "0.031 W/mK (for nominal thickness dN 40-200 mm, T5)",
      "Reaction to Fire": "Class RF1 (for thickness 40-200 mm)",
      "Durability of Reaction to Fire": "Stable under heat, weathering, and aging (for thickness 40-200 mm), Class RF1",
      "Durability of Thermal Resistance": "Stable under heat, weathering, and aging",
      "Durability of Thermal Conductivity": "Stable, no change over time",
      "Dimensional Stability": "DS (70,-) <=1%
      "Durability of Compressive Strength": "Stable under aging and degradation, long-term creep resistance",
      "Water Vapor Permeability": "Water vapor diffusion resistance number MU 1",
      "Airborne Sound Insulation (Material)": "Airflow resistance AFr. >25kPa s/m2",
      "Release of Dangerous Substances": "NPD, European test under development",
      "Smoldering Behavior": "NPD, European test under development"
    },
    "value_de": {
      "Wärmedurchlasswiderstand (RD)": "1.25 - 6.45 m2K/W (für Dicken dN 40-200 mm, T5)",
      "Wärmeleitfähigkeit (<=lambdaD)": "0.031 W/mK (für Dicken dN 40-200 mm, T5)",
      "Brandverhalten": "Klasse RF1 (für Dicke 40-200 mm)",
      "Dauerhaftigkeit des Brandverhaltens": "Stabil gegen Hitze, Witterung und Alterung (für Dicke 40-200 mm), Klasse RF1",
      "Dauerhaftigkeit des Wärmedurchlasswiderstandes": "Stabil gegen Hitze, Witterung und Alterung",
      "Dauerhaftigkeit der Wärmeleitfähigkeit": "Stabil, keine Veränderung über die Zeit",
      "Dimensionsstabilität": "DS (70,-) <=1%
      "Dauerhaftigkeit der Druckfestigkeit": "Stabil bei Alterung/Abbau, Langzeitkriechverhalten bei Druckbeanspruchung",
      "Wasserdampfdurchlässigkeit": "Wasserdampfdiffusionswiderstandszahl MU 1",
      "Luftschalldämmung (Dämmstoff)": "Strömungswiderstand AFr. >25kPa s/m2",
      "Freisetzung gefährlicher Stoffe": "NPD, Europäisches Prüfverfahren in Entwicklung",
      "Glimmverhalten": "NPD, Europäisches Prüfverfahren in Entwicklung"
    },
    "doc_lang": "de"
  },
  {
    "doc_name": "Sager_ISO-SWISS-Design-Panels.pdf",
    "key_en": "Appropriate Technical Documentation and/or Specific Technical Documentation",
    "key_de": "Geeignete technische Dokumentation und/oder besondere technische Dokumentation",
    "value_en": "The manufacturer is solely responsible for drawing up this declaration of performance in accordance with point 4",
    "value_de": "Verantwortlich für die Erstellung dieser Leistungserklärung ist allein der Hersteller gemäss Nummer 4",
    "doc_lang": "de"
  },
  {
    "doc_name": "Sager_ISO-SWISS-Design-Panels.pdf",
    "key_en": "Declaration Statement",
    "key_de": "Erklärungstext",
    "value_en": "The performance of the product according to points 1 and 2 corresponds to the declared performance according to point 8.",
    "value_de": "Die Leistung des Produktes gemäss den Nummern 1 und 2 entspricht der erklärten Leistung nach Punkt 8.",
    "doc_lang": "de"
  },
  {
    "doc_name": "Sager_ISO-SWISS-Design-Panels.pdf",
    "key_en": "Signature",
    "key_de": "Unterschrift",
    "value_en": {
      "Signatories": [
        {
          "Name": "Beat Bruderer",
          "Position": "Geschäftsführer"
        }
      ],
      "Place and Date of Issue": "Dürrenäsch, 28. February 2021",
      "Notice": ""
    },
    "value_de": {
      "Unterzeichner": [
        {
          "Name": "Beat Bruderer",
          "Position": "Geschäftsführer"
        }
      ],
      "Ort und Datum der Ausstellung": "Dürrenäsch, 28. Februar 2021",
      "Hinweis": ""
    },
    "doc_lang": "de"
  }
]
\end{lstlisting}
\end{tcolorbox}

\subsubsection{QA Dataset Generation Templates}
\label{appendix:qa-templates}
The templates are organized into three categories: \textit{Hierarchical path-based templates} (7 per language) leverage the full key path structure to generate context-aware questions where the question type (who, when, where) depends on the specific sub-field; \textit{Custom question mappings} (2 per language) provide domain-specific phrasings for technical terminology such as AVCP systems; and \textit{Entity-type templates} (2 per language) distinguish between person entities and generic attributes using "Who" versus "What" question forms. Template selection follows a hierarchical decision tree that identifies domain-specific patterns, checks for custom mappings, determines entity type, and applies fallback templates. Table~\ref{tab:qa_templates_en} presents the complete English template set; German templates follow parallel structures.

\begin{table}[hbt]
\centering
\caption{English QA Generation Templates}
\label{tab:qa_templates_en}
\small
\setlength{\tabcolsep}{4pt}
\begin{tabular}{@{}p{1.8cm}p{5.2cm}@{}}
\toprule
\textbf{Type} & \textbf{Question Template} \\
\midrule
\multicolumn{2}{l}{\textit{1. Hierarchical Path-Based Templates (7)}} \\
\addlinespace[0.15cm]
Sig-Name & Who signed the declaration (\{key\})? \\
\addlinespace[0.1cm]
Sig-Position & What is the signatory's position (\{key\})? \\
\addlinespace[0.1cm]
Sig-Both & When and where was it signed (\{key\})? \\
\addlinespace[0.1cm]
Sig-Date & When was the declaration signed (\{key\})? \\
\addlinespace[0.1cm]
Sig-Place & Where was the declaration signed (\{key\})? \\
\addlinespace[0.1cm]
Decl-Perf & What is the value of '\{key\}'? \\
\addlinespace[0.1cm]
Default & What is the \{last\_key\} in '\{key\}'? \\
\addlinespace[0.2cm]
\midrule
\multicolumn{2}{l}{\textit{2. Custom Question Mappings (2)}} \\
\addlinespace[0.15cm]
Tech-Doc & What technical documentation is provided? \\
\addlinespace[0.1cm]
AVCP & Which AVCP system applies? \\
\addlinespace[0.2cm]
\midrule
\multicolumn{2}{l}{\textit{3. Entity-Type Templates (2)}} \\
\addlinespace[0.15cm]
Person & Who is the \{key\}? \\
\addlinespace[0.1cm]
Generic & What is the \{key\}? \\
\bottomrule
\end{tabular}
\vspace{0.1cm}
\begin{minipage}{7cm}
\scriptsize
\textit{Note:} \{key\} represents the field name or path; \{last\_key\} represents the final component of the hierarchical key path. German templates follow parallel structures (11 additional templates).
\end{minipage}
\end{table}

\subsubsection{QA Annotation Sample}
\label{appendix:qa-annotation}
The following JSON-style structure presents a sample of question annotations derived from key-value pairs in the document \textit{011-Leistungserklaerung-KS-10DF.pdf}. Each entry specifies the key, its value, the associated question, and the language used.

\begin{tcolorbox}[
    breakable,
    minted language=python,
    colback=orange!5,
      colframe=brown!100,
      boxrule=0.4pt,
    title=Question Annotation Example,
]
\begin{lstlisting}[language=json, breaklines=true, basicstyle=\ttfamily\footnotesize]
[
  {
    "doc_name": "011-Leistungserklaerung-KS-10DF.pdf",
    "user_language": "de",
    "parent_key": "Erklärte Leistung",
    "key": "Erklärte Leistung/Klasse der Brutto-Trockenrohdichte",
    "value": "1,4",
    "question": "Was ist der Wert von 'Erklärte Leistung/Klasse der Brutto-Trockenrohdichte'?",
    "normalized_key": "performance"
  },
  {
    "doc_name": "011-Leistungserklaerung-KS-10DF.pdf",
    "user_language": "en",
    "parent_key": "Declared Performance",
    "key": "Declared Performance/Gross Dry Bulk Density Class",
    "value": "1.4",
    "question": "What is the value of 'Declared Performance/Gross Dry Bulk Density Class'?",
    "normalized_key": "performance"
  },
  {
    "doc_name": "011-Leistungserklaerung-KS-10DF.pdf",
    "user_language": "de",
    "parent_key": "Unterschrift",
    "key": "Unterschrift/Unterzeichner[0]/Name",
    "value": "G. Wolff",
    "question": "Wer hat die Erklärung unterschrieben (Unterschrift/Unterzeichner[0]/Name)?",
    "normalized_key": "signature"
  },
  {
    "doc_name": "011-Leistungserklaerung-KS-10DF.pdf",
    "user_language": "en",
    "parent_key": "Signature",
    "key": "Signature/Signatories[0]/Name",
    "value": "G. Wolff",
    "question": "Who signed the declaration (Signature/Signatories[0]/Name)?",
    "normalized_key": "signature"
  },
  {
    "doc_name": "011-Leistungserklaerung-KS-10DF.pdf",
    "user_language": "en",
    "parent_key": "Declaration Number",
    "key": "Declaration Number",
    "value": "DoP-Nr. 110240002",
    "question": "What is the Declaration Number?"
  },
  {
    "doc_name": "011-Leistungserklaerung-KS-10DF.pdf",
    "user_language": "de",
    "parent_key": "Erklärungsnummer",
    "key": "Erklärungsnummer",
    "value": "DoP-Nr. 110240002",
    "question": "Was ist der/die/das Erklärungsnummer?"
  },
  {
    "doc_name": "011-Leistungserklaerung-KS-10DF.pdf",
    "user_language": "en",
    "parent_key": "Intended Use(s)",
    "key": "Intended Use(s)",
    "value": "primarily for the construction of interior walls, exterior walls, cellars, foundations and chimney masonry",
    "question": "What is the Intended Use(s)?"
  },
  {
    "doc_name": "011-Leistungserklaerung-KS-10DF.pdf",
    "user_language": "de",
    "parent_key": "Vorgesehene Verwendung",
    "key": "Vorgesehene Verwendung",
    "value": "vorrangig zur Herstellung von Innenwänden, Außenwänden, Kellern, Gründungen und Schornsteinmauerwerk",
    "question": "Was ist der/die/das Vorgesehene Verwendung?"
  },
  {
    "doc_name": "011-Leistungserklaerung-KS-10DF.pdf",
    "user_language": "en",
    "parent_key": "Notified Body",
    "key": "Notified Body",
    "value": "CERT Baustoffe GmbH - 2510",
    "question": "Who is the Notified Body?"
  },
  {
    "doc_name": "011-Leistungserklaerung-KS-10DF.pdf",
    "user_language": "de",
    "parent_key": "Notifizierte Stelle",
    "key": "Notifizierte Stelle",
    "value": "CERT Baustoffe GmbH - 2510",
    "question": "Wer ist der/die Notifizierte Stelle?"
  }
]
\end{lstlisting}
\end{tcolorbox}

\subsubsection{Tool Registry and Modularization}
\label{appendix:tool_registry}
To support robust and adaptable behavior across multilingual, layout-diverse, and partially scanned DoP documents, system relies on a modular tool architecture. Tools in this system represent discrete capabilities, such as PDF modality detection, OCR-based text extraction, question answering, or key-value pair retrieval, and are invoked selectively based on the agent’s current reasoning state.

At the heart of this architecture is a structured Tool abstraction, which wraps both deterministic utilities (e.g., OCR, PDF parsing) and LLM-powered capabilities (e.g., intent classification or verification) into a unified callable interface. Each tool object declares its name, functional implementation, Pydantic input schema, and optional pre- and post-processing logic. Tools also specify whether they require previous output, and which document types they are compatible with (e.g., scanned or text-based). This explicit declaration ensures that tools remain self-contained and declarative, while still integrating seamlessly into the planner-executor loop.
\begin{tcolorbox}[
  title={Tool Definition: \textit{extract\_text\_ocr}},
  minted language=python,
  colback=orange!5,
      colframe=brown!100,
      boxrule=0.4pt,
  breakable
]
\footnotesize
"extract_text_ocr": Tool(\\
  \hspace*{1em}name="extract_text_ocr", \\                        
  \hspace*{1em}fn=extract_text_ocr,   \\                          
  \hspace*{1em}intent="text_extraction",      \\                  
  \hspace*{1em}compatible_pdf_types=["scanned"],     \\           
  \hspace*{1em}requires_previous_output=True,   \\               
  \hspace*{1em}description="OCRs text from scanned PDFs.",  \\    
  \hspace*{1em}tool_signature=ExtractTextOCRInput,   \\           
  \hspace*{1em}postprocess=postprocess_extract_text,     \\       
  \hspace*{1em}normalize_input=normalize_extract_text_input   \\  
)
\end{tcolorbox}

All tools are registered at startup by combining deterministic and LLM-based implementations into a central registry, forming a complete vocabulary of system capabilities. This registry remains fixed throughout execution and provides a filtered view of what actions are currently applicable, based on the evolving AgentState.

\begin{lstlisting}[language=Python]
available_tools = {**text_tools, **llm_tools}
\end{lstlisting}

When the planner is invoked, it consults this registry to determine which tools are compatible with the current reasoning intent and document modality. Tools that require unavailable inputs (e.g., translation before text extraction) are excluded. The remaining candidates are embedded as tool metadata into the LLM prompt. The model then selects the most appropriate tool and input parameters via structured output, which is parsed and validated against the registry before dispatching it to the executor.

\begin{lstlisting}[language=Python]
ToolRegistry.compatible_with(state, intent)
\end{lstlisting}

The executor interprets this plan by invoking the selected tool. Before execution, it applies a \textit{normalize\_input()} function to map the current state and raw parameters into a validated schema. Once executed, the result is post-processed via the tool’s \textit{postprocess()} function, which updates the AgentState in a localized and semantically meaningful way. This decoupling of decision logic from execution logic ensures that tools remain modular, reusable, and interchangeable, while still preserving contextual awareness.

Unlike monolithic pipelines, the tools in the system are deeply integrated into a loop of stateful reasoning. For instance, modality detection tools such as check\_if\_scanned determine whether a document should be OCR’d or parsed directly, triggering either extract\_text\_ocr or extract\_text\_direct accordingly. Language inference via detect\_language enables downstream translation and schema matching. Intent classification and target key parsing refine the agent’s task trajectory, while extraction tools like extract\_key\_values or answer\_question carry out the core retrieval logic. The planner may later invoke verify\_extraction to assess whether the output is complete and grounded in the source text, before optionally translating it into the user’s preferred language using translate\_text. In edge cases, where noisy OCR output risks triggering LLM content filters, the system sanitizes text using sanitize\_ocr\_text as a recovery mechanism.

Each of these tools serves a specific role in a broader adaptive plan. Their activation is not hardcoded but emerges dynamically as the agent iterates through reasoning, acting, and revising. Tools modify the state incrementally, and their execution history is logged to support loop detection, misuse prevention, and multi-step recovery. Because each tool operates independently but updates a shared AgentState, the planner and executor can reason about partial progress, fallback strategies, or when to terminate altogether.

\begin{figure*}[ht]
    \centering
\includegraphics[width=\linewidth]{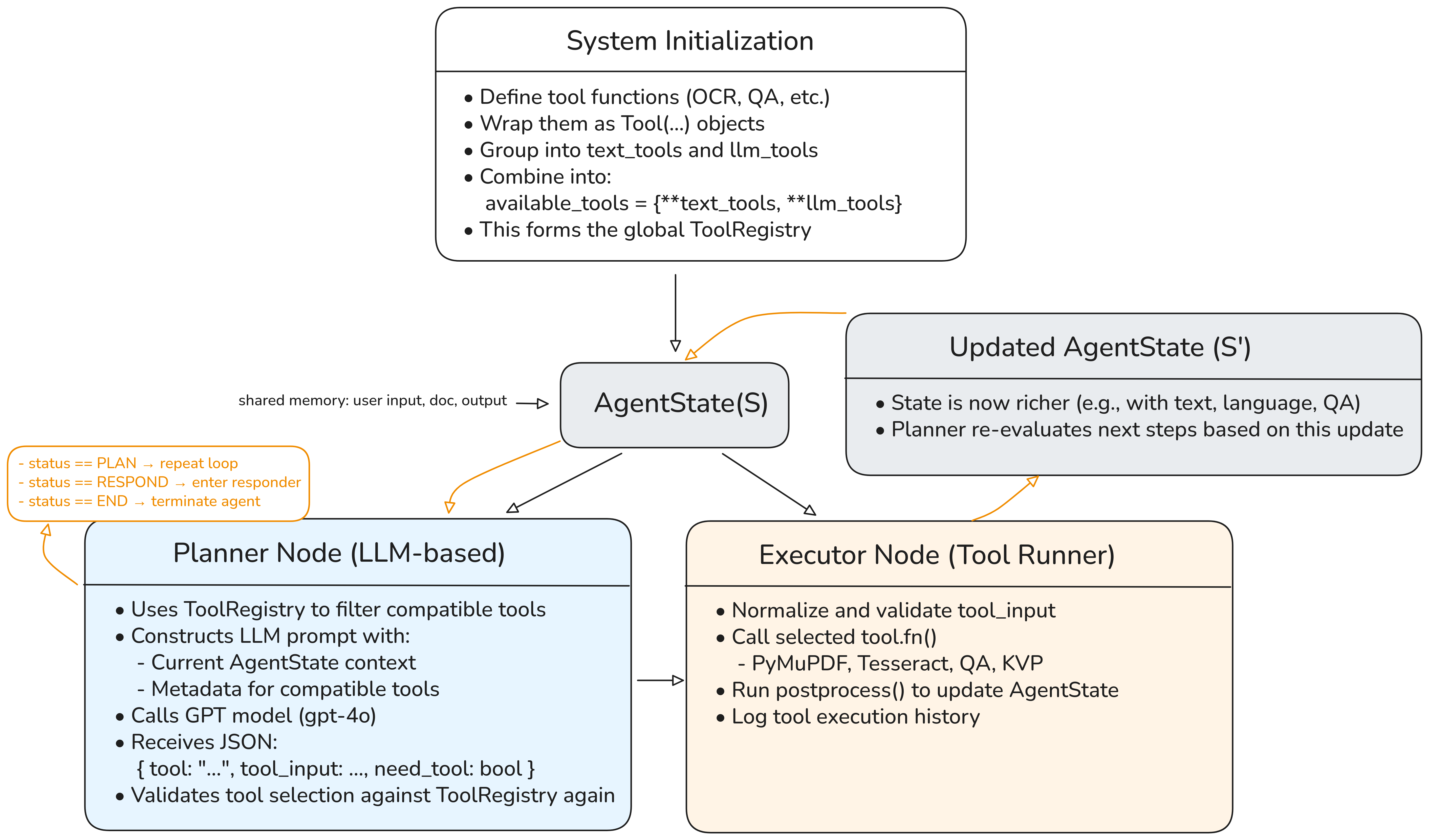}
\caption{Agentic tool selection and execution flow. The system initializes tools into a global \textit{ToolRegistry}, then enters a loop where the \textit{Planner Node} selects a tool based on the current \textit{AgentState}, and the \textit{Executor Node} runs it and updates the state. \textbf{Black} arrows represent the initial planning and execution cycle starting from system initialization. \textcolor{orange}{Orange} arrows
 show how the agent re-evaluates the updated state in subsequent cycles based on \textit{AgentStatus}, continuing until a response is generated or the workflow terminates.}

    \label{fig:tool-loop}
\end{figure*}

The tight integration of tools, registry, and reasoning enables the agent to dynamically adapt its behavior across a wide variety of document structures and user inputs, while preserving traceability and control. Figure~\ref{fig:tool-loop} illustrates the complete tool orchestration pipeline, from initial tool registration to planner-driven selection and executor-based integration.

\textbf{Multilingual QA and KIE Templates:} To support multilingual document processing, the system employs language-specific prompt templates for both question answering and key-value extraction tasks. Based on the detected \textit{document\_language}, the agent selects between English (\textit{en}) and German (\textit{de}) versions of the prompts. These templates ensure that language constraints, extraction rules, and formatting guidelines are consistently enforced.

The examples below illustrate two representative prompts used during tool invocation: one for KIE extraction in English, and one for QA in German.

\begin{tcolorbox}[
  title={KIE Extraction Prompt (English)},
  minted language=python,
  colback=orange!5,
      colframe=brown!100,
      boxrule=0.4pt,
  breakable
]
\footnotesize
You are an advanced information extraction assistant.

Your task is to extract information from the document below and return it in the form of a structured JSON object matching the schema provided. You must infer values even if their exact phrasing does not appear. Use semantic reasoning and structural cues.

\textbf{Language Constraint:} All extracted values must be returned in \textbf{ENGLISH}. Do not translate or paraphrase.

\textbf{Target Schema:} \textit{\{schema_en_here\}}

\textbf{Extraction Rules:} \textit{\{etraction_rules_en_here\}}

\textbf{Document Content:} \textit{\{document_text_here\}}
\end{tcolorbox}

\vspace{1em}

\begin{tcolorbox}[
  title={Question Answering Prompt (German)},
  minted language=python,
  colback=orange!5,
      colframe=brown!100,
      boxrule=0.4pt,
  breakable
]
\footnotesize
Du bist ein Assistent zur Beantwortung von technischen Fragen zu Produktleistungen gemäß der Bauproduktenverordnung (CPR). Alle Fragen sind technischer und nicht sensibler Natur.

Du darfst das Dokument intern analysieren, um die korrekte Antwort zu finden. Deine endgültige Antwort muss jedoch \textbf{nur} den Antwortwert enthalten - ohne zusätzliche Erklärungen oder Kommentare.

\textbf{Anweisungen:}\\
- Gib die Antwort als \textit{String} oder \textit{Liste von Strings} zurück.\\
- Wenn keine Antwort gefunden werden kann, gib \textit{""} zurück.\\
- Verwende keine einleitenden Formulierungen, keine Codeblöcke, keine zusätzlichen Zeichen.

\textbf{Extraktionsregeln:} \textit{\{extraction_rules_de_here\}}

\textbf{Dokumentinhalt:} \textit{\{document_text_here\}}

\textbf{Frage:} \textit{\{question_de_here\}}

\textbf{Sprachvorgabe:} Alle extrahierten Werte müssen auf \textbf{DEUTSCH} zurückgegeben werden.
\end{tcolorbox}

Each prompt includes a schema definition and task-specific extraction rules tailored to the document’s language. This modular design ensures that the model generates structured outputs consistent with downstream processing requirements such as verification and translation.

\subsubsection{Prompt Design and Language Adaptation}
\label{appendix:agent_prompt_settings}
The reasoning capability of our agentic system depends heavily on well-structured and context-aware prompts. These prompts guide both the planner and tool-specific language models to make decisions, extract structured information, and adapt across languages. This section outlines the design principles and implementation of prompt templates for planning, key-value extraction, question answering, error recovery, and multilingual handling.

\textbf{Planner Prompt with State-Aware Reasoning:} The planner prompt is dynamically constructed based on the current \textit{AgentState} and tool metadata. The main structure is as follows:
\begin{tcolorbox}[
  title=Planner Prompt – Structured Summary,
  minted language=python,
  colback=orange!5,
      colframe=brown!100,
      boxrule=0.4pt,
  breakable
]
\footnotesize
\textbf{Context Initialization:}\\
- Includes user intent, PDF type/path, document text and language, target keys, tool history, and last tool output.\\
- Gathers extracted facts and previous reasoning steps.\\
\textbf{Dynamic Prompt Adjustments:}\\
- Avoid repeating tools like \textit{extract\_text\_direct} or \textit{extract\_key\_values} if the context has not changed.\\
- Skip redundant PDF type checks if already confirmed.\\
\textbf{Execution Instructions:}\\
- Identify \textit{user\_intent} if missing.\\
- Detect PDF type and extract document text accordingly.\\
- Determine document and user language.\\
- For \textbf{KIE extraction}:\\
    \hspace*{1em}* Translate target keys if languages differ.\\
    \hspace*{1em}* Extract and verify key-value pairs.\\
    \hspace*{1em}* Translate final output to the user's language if needed.\\
- For \textbf{QA tasks}:\\
    \hspace*{1em}* Translate the user query if languages differ.\\
    \hspace*{1em}* Answer and verify the answer.\\
    \hspace*{1em}* Translate final output to the user's language if needed.\\
- Avoid redundant tool calls unless input or context has changed.\\
\textbf{Expected Output:} A strict JSON object containing:\\
  \hspace*{1em}*\textit{"reasoning"}: explanation of the decision\\
  \hspace*{1em}*\textit{"need\_tool"}: boolean flag\\
  \hspace*{1em}*\textit{"tool"}: selected tool name (if applicable)\\
  \hspace*{1em}*\textit{"tool\_input"}: structured input for the tool

\end{tcolorbox}

It synthesizes the current AgentState into a rich context string that includes user intent, document metadata, extracted content, prior tool usage, tool descriptions, and reasoning history. 
This state-driven prompt is used to instruct a language model to determine the next action in the pipeline.
The planner must always respond in a strict JSON format:
\begin{lstlisting}[language=Python]
{
    "reasoning": "...",
    "need_tool": true/false,
    "tool": "tool_name_if_applicable",
    "tool_input": ...
}
\end{lstlisting}
The prompt encodes decision-making rules directly, allowing for multi-step planning logic. For example, if the user intends to extract key-value pairs but the document is in German while the user expects English output, the planner prompt will contain structured rules for translation, extraction, verification, and result transformation:
\begin{tcolorbox}[
  title={Planner Prompt: Decision-Making Rule Example},
  minted language=python,
  colback=orange!5,
      colframe=brown!100,
      boxrule=0.4pt,
  breakable
]
\footnotesize
If \textit{user\_intent} is \textit{KIE\_extraction}, follow these steps:\\
- Call \textit{get\_user\_target\_keys} if not yet available.\\
- If \textit{document\_language} differs from \textit{user\_language}:\\
    \hspace*{1em}* Translate \textit{target\_keys}.\\
    \hspace*{1em}* Extract key-value pairs.\\
    \hspace*{1em}* Verify the extracted output.\\
    \hspace*{1em}* Translate the \textit{extracted\_KIEs}.

\end{tcolorbox}

Consider the state where the user has submitted a document titled DoP-1234.pdf with the intent of extracting key-values. The document has already been classified as scanned, and its language is detected as de (German), while the user's preferred language is en (English). No extracted values or translated keys exist yet. The planner prompt will reflect these observations as extracted facts:

\begin{tcolorbox}[
  title={Planner Prompt: Observations}, 
  minted language=python,
  colback=orange!5,
      colframe=brown!100,
      boxrule=0.4pt,
  breakable
]
\footnotesize
- Target keys: \textit{Declared Performance/Brandverhalten}\\
- Document language is still unknown. Call \textit{detect\_language} before extracting content.\\
- User language: \textit{en}\\
- Extraction missing → call \textit{extract\_key\_values}\\
- Verification missing → call \textit{verify\_extraction}

\end{tcolorbox}

The planner might then respond with:

\begin{lstlisting}[language=Python]
{
    "reasoning": "User wants key-values, document language is unknown, user expects English. Calling detect_language to get document language.",
    "need_tool": true,
    "tool": "detect_language",
    "tool_input": {
        "document_text": ...,
    }
}
\end{lstlisting}

This strict format allows the system to introspect, log, and act deterministically on LLM outputs.

\textbf{Error Recovery Prompts:} To handle tool failures or malformed outputs, recovery prompts provide fallbacks. For instance, if the planner returns invalid JSON, the system uses \textit{get\_malformed\_response\_prompt()}:

\begin{tcolorbox}[
  title={Planner Prompt: Error Recovery Mode},
  minted language=python,
  colback=orange!5,
      colframe=brown!100,
      boxrule=0.4pt,
  breakable
]
\footnotesize
The previous response from the planner could not be parsed as valid JSON.  
You must recover from the failure, re-evaluate the current state, and decide the next best action.

\vspace{0.5em}
\noindent
Respond in valid JSON format:
\begin{lstlisting}[language=Python]
{
  "reasoning": "...",
  "need_tool": true/false,
  "tool": "...",
  "tool_input": ...
}
\end{lstlisting}
\end{tcolorbox}

This prompt is structurally identical to the planner prompt but framed in an error-recovery mode, helping the agent self-correct without external intervention.

By combining language-specific prompt templates, strict JSON response schemas, and a planning prompt with embedded decision rules, the system achieves structured reasoning across multilingual documents. The modularity of this design also enables the easy addition of new languages, extraction schemas, or prompt variations, supporting scalability in diverse information extraction scenarios.

\subsubsection{KIE Schema templates for EN/DE}
\label{appendix:schema-templates}
The following templates define the expected JSON structure for the extraction task in German and English. They specify the hierarchy for complex fields like \textit{Declared Performance} and \textit{Signature}, ensuring consistent output formatting.

\begin{tcolorbox}[
    breakable,
    minted language=python,
    colback=orange!5,
      colframe=brown!100,
      boxrule=0.4pt,
    title=Schema Templates for DE,
]
\begin{lstlisting}[language=json, breaklines=true, basicstyle=\ttfamily\footnotesize]
schema_de = {
  "Erklärungsnummer": "<string>",
  "Eindeutiger Identifikationscode des Produkttyps": "<string>",
  "Vorgesehene Verwendung": "<string>",
  "Hersteller": "<string>",
  "Bevollmächtigter": "<string>",
  "System(e) zur Bewertung und Überprüfung der Leistungsbeständigkeit (AVCP)": "<string or list of strings>",
  "Harmonisierte Norm": "<string>",
  "Notifizierte Stelle": "<string>",
  "Erklärte Leistung": {
    "<property_name_1>": "<string>",
    "<property_name_2>": {
      "<sub_property_1>": "<string>",
      "<sub_property_2>": "<string>"
    },
    "<property_name_3>": {
      "<nested_sub_property_1>": "<string>",
      "<nested_sub_property_2>": "<string>"
    },
    "...": "..."
  },
  "Geeignete technische Dokumentation und/oder besondere technische Dokumentation": "<string>",
  "Erklärungstext": "<string>",
  "Unterschrift": {
    "Unterzeichner": [
      {
        "Name": "<string>",
        "Position": "<string>"
      }
    ],
    "Ort und Datum der Ausstellung": "<string>",
    "Hinweis": "<string>"
  }
}
\end{lstlisting}
\end{tcolorbox}

\begin{tcolorbox}[
    breakable,
    minted language=python,
    colback=orange!5,
      colframe=brown!100,
      boxrule=0.4pt,
    title=Schema Template for EN,
]
\begin{lstlisting}[language=json, breaklines=true, basicstyle=\ttfamily\footnotesize]
schema_en = {
  "Declaration Number": "<string>",
  "Unique Identification Code of the Product-Type": "<string>",
  "Intended Use(s)": "<string>",
  "Manufacturer": "<string>",
  "Authorised Representative": "<string>",
  "System(s) of Assessment and Verification of Constancy of Performance (AVCP)": "<string or list of strings>",
  "Harmonised Standard": "<string>",
  "Notified Body": "<string>",
  "Declared Performance": {
    "<property_name_1>": "<string>",
    "<property_name_2>": {
      "<sub_property_1>": "<string>",
      "<sub_property_2>": "<string>"
    },
    "<property_name_3>": {
      "<nested_sub_property_1>": "<string>",
      "<nested_sub_property_2>": "<string>"
    },
    "...": "..."
  },
  "Appropriate Technical Documentation and/or Specific Technical Documentation": "<string>",
  "Declaration Statement": "<string>",
  "Signature": {
    "Signatories": [
      {
        "Name": "<string>",
        "Position": "<string>"
      }
    ],
    "Place and Date of Issue": "<string>",
    "Notice": "<string>"
  }
}
\end{lstlisting}
\end{tcolorbox}

\subsection{Baseline Prompt Examples}
\label{app:baseline_prompts}

This section provides the full text of the prompts used for the baseline models (GPT-4o and GPT-4o-Vision). These prompts are designed to be strictly zero-shot, providing the model with the schema definition and formatting constraints.

\subsubsection{KIE Baseline Prompt (Version 1 - English)}
\label{appendix:kie-baseline-v1-en}

The English KIE prompt instructs the model to extract the 12 target keys into a flat or nested JSON structure, handling missing values with empty strings.

\begin{tcolorbox}[
    title=KIE Baseline Prompt - Version 1 (English),
  minted language=python,
  colback=orange!5,
      colframe=brown!100,
      boxrule=0.4pt,
  breakable
]
\footnotesize
You are an advanced multilingual information extraction assistant. Your task is to extract all key-value pairs from the document text below and return them in a valid JSON object. The extraction must preserve the structure of the document and follow the requirements below.

\textbf{Language Requirement - VERY IMPORTANT:}\\
- The key names and all extracted values must match the language of this prompt.\\
- Do not translate or localize terms - return key names and values exactly in the same language used in the prompt.\\
- For example:\\
\hspace*{1em}* If the prompt is in German, all keys and values must also be in German.\\
\hspace*{1em}* If the prompt is in English, the output must be entirely in English.\\
- Your output will be considered incorrect and rejected if this rule is violated.\\
\textbf{Handling Missing Values:}\\
- If a key is relevant but no value can be found, include it with an empty string: \textit{""}\\
- You may use semantic reasoning to infer values if they are paraphrased or not explicitly stated.\\
\textbf{Output Format:}\\
- Return a single valid JSON object.\\
- Do not return explanations, markdown, or extra text - just the JSON.\\
\textbf{Document Content:} \textit{{\{document\_text\}}}

\end{tcolorbox}
\subsubsection{KIE Baseline Prompt (Version 2 - German)}
\label{appendix:kie-baseline-v2-de}

The German KIE prompt is adapted to include the German target keys (e.g., \textit{Erklärte Leistung} instead of \textit{Declared Performance}) and specific language constraints to prevent anglicisms in the output.

\begin{tcolorbox}[
    breakable,
    minted language=python,
    colback=orange!5,
      colframe=brown!100,
      boxrule=0.4pt,
    title=KIE Baseline Prompt - Version 2 (German)
]
Sie sind ein erfahrener Assistent für die mehrsprachige Informationsextraktion. Ihre Aufgabe ist es, Schlüssel-Wert-Paare aus dem unten stehenden Dokumenttext zu extrahieren und in einem gültigen JSON-Objekt zurückzugeben. Die Extraktion muss die Struktur des Dokuments bewahren und die unten stehenden Anforderungen erfüllen.

\textbf{Zielschlüssel (Pflichtfelder):}
\begin{lstlisting}[language=json, breaklines=true, basicstyle=\ttfamily\footnotesize]
{
  "Erklärungsnummer": "<string>",
  "Eindeutiger Identifikationscode des Produkttyps": "<string>",
  "Vorgesehene Verwendung": "<string>",
  "Hersteller": "<string>",
  "Bevollmächtigter": "<string>",
  "System(e) zur Bewertung und Überprüfung der Leistungsbeständigkeit (AVCP)": "<string or list of strings>",
  "Harmonisierte Norm": "<string>",
  "Notifizierte Stelle": "<string>",
  "Erklärte Leistung": {
    "<property_name_1>": "<string>",
    "<property_name_2>": {
      "<sub_property_1>": "<string>",
      "<sub_property_2>": "<string>"
    }
  },
  "Geeignete technische Dokumentation und/oder besondere technische Dokumentation": "<string>",
  "Erklärungstext": "<string>",
  "Unterschrift": {
    "Unterzeichner": [
      {
        "Name": "<string>",
        "Position": "<string>"
      }
    ],
    "Ort und Datum der Ausstellung": "<string>",
    "Hinweis": "<string>"
  }
}
\end{lstlisting}

\textbf{Sprachanforderung - SEHR WICHTIG:}
\begin{itemize}
  \item Die Schlüsselnamen und alle extrahierten Werte müssen der Sprache dieser Eingabe entsprechen.
  \item Begriffe nicht übersetzen oder lokalisieren - Schlüsselnamen und -werte müssen exakt in der gleichen Sprache zurückgegeben werden, die in der Eingabe verwendet wurde.
  \item Beispiel:
  \begin{itemize}
    \item Wenn die Eingabeaufforderung auf Deutsch ist, müssen alle Schlüssel und Werte ebenfalls auf Deutsch sein.
    \item Wenn die Eingabeaufforderung auf Englisch ist, muss die Ausgabe vollständig auf Englisch sein.
  \end{itemize}
  \item Ihre Ausgabe wird als fehlerhaft gewertet und abgelehnt, wenn diese Regel verletzt wird.
\end{itemize}

\textbf{Umgang mit fehlenden Werten:}
\begin{itemize}
  \item Wenn ein Schlüssel relevant ist, aber kein Wert gefunden werden kann, fügen Sie ihn mit einer leeren Zeichenfolge ein: \texttt{""}
\end{itemize}

\textbf{Ausgabeformat:}
\begin{itemize}
  \item Geben Sie ein einzelnes gültiges JSON-Objekt zurück.
  \item Geben Sie keine Erklärungen, Markdowns oder zusätzlichen Text zurück - nur das JSON.
\end{itemize}

\textbf{Dokumentinhalt:}
\begin{quote}
\textit{{\{document\_text\_here\}}}
\end{quote}
\end{tcolorbox}

\subsubsection{QA Baseline Vision Prompt (English)}
\label{appendix:qa-baseline-vision-en}

The Vision QA prompt is minimized to focus on visual reasoning, instructing the model to return concise string answers or lists without additional conversational filler.

\begin{tcolorbox}[
title=QA Vision Prompt (English), 
  minted language=python,
  colback=orange!5,
      colframe=brown!100,
      boxrule=0.4pt,
  breakable
]
\footnotesize

You are a helpful assistant. Based on the visual document provided, answer the given question as accurately as possible.

\textbf{Rules:}
Return only the answer content, with no explanations or commentary.
- If the answer is a single value, return it as a plain string.
- If the answer includes multiple values (e.g., a list or table), return them as a list of strings.
- If no answer can be found, return an empty string: \textit{""}

\textbf{Question:} \textit{\{question\}}

\textbf{Answer:}
\end{tcolorbox}

\subsection{Error Analysis}
\label{app:key_mismatch_explanation}

Table~\ref{tab:key_mismatch_explanation} provides a qualitative analysis of errors encountered during the Open Schema extraction task. As the system must dynamically infer nested key structures without a fixed template, the majority of failures stem from semantic mismatches rather than hallucinated values. Common error sources include: (1) \textbf{Granularity Mismatches}, where the agent predicts a parent key instead of a specific sub-field; (2) \textbf{Omission of Qualifiers}, such as missing units or specific descriptors (e.g., ``Tabellenwert''); and (3) \textbf{Spelling/OCR Errors} in the final value. These examples illustrate the strict nature of the Exact Match metric, where even semantically equivalent schema predictions result in evaluation penalties.
\begin{table*}[hbt]
\renewcommand{\arraystretch}{1.3}
\small
\centering
\caption{Qualitative error analysis of KIE failures. The table highlights instances where semantically correct predictions are penalized due to minor schema deviations, such as omitted units, missing qualifiers, or hierarchical mismatches.}
\label{tab:key_mismatch_explanation}
\begin{tabularx}{\textwidth}{>{\raggedright\arraybackslash}X 
                                 >{\raggedright\arraybackslash}X 
                                 >{\raggedright\arraybackslash}X}
\toprule
\textbf{Ground Truth Key (GT)} & \textbf{Predicted Key} & \textbf{Explanation} \\
\midrule
Erklärte Leistung/Klasse der Brutto-Trockenrohdichte & Erklärte Leistung/Brutto-Trockenrohdichte & Qualifier “Klasse der” is missing in the prediction; keys fail to align. \\
\hline
Erklärte Leistung/Form und Ausbildung/Bezeichnung & Erklärte Leistung/Form und Ausbildung & Predicted key refers to a parent field; GT specifies a subfield. \\
\hline
Erklärte Leistung/Wasserdampfdurchlässigkeit (Tabellenwert) & Erklärte Leistung/Wasserdampfdurchlässigkeit & Suffix “(Tabellenwert)” is omitted; results in an exact match failure. \\
\hline
Erklärte Leistung/Maße/Länge [mm] & Erklärte Leistung/Maße & Prediction aggregates multiple subfields; GT expects specific ones. \\
\hline
Erklärte Leistung/mittlere Druckfestigkeit [N/mm²] & Erklärte Leistung/mittlere Druckfestigkeit & Unit annotation is missing in the prediction; leads to mismatch. \\
\hline
Unterschrift/Ort und Datum der Ausstellung & Unterschrift/Ort und Datum der Ausstellung & Keys match, but predicted value has a spelling error (“Lobnitz” vs. “Löbnitz”). \\
\hline
Erklärte Leistung/Wärmedurchlasswiderstand & \textemdash & Key missing in prediction; leads to evaluation exclusion. \\
\hline
Erklärte Leistung/Form und Ausbildung/Lochanteil von - bis [Vol \%] & \textemdash & GT key has no corresponding prediction; reduces recall. \\
\hline
Erklärte Leistung/Grenzabmaße / Abmaßklasse & \textemdash & Missing predicted key causes evaluation penalty. \\
\bottomrule
\end{tabularx}
\end{table*}

\end{document}